\documentclass[runningheads]{llncs}
\usepackage{graphicx}
\usepackage{amsmath,amssymb}
\usepackage{color}
\usepackage{booktabs}
\usepackage{longtable} 
\usepackage{url}

\usepackage{subcaption}
\newcommand\etal{\textit{et al. }}

\begin{document}
\pagestyle{headings}
\mainmatter

\title{Failure Detection for Facial Landmark Detectors} 

\author{Andreas Steger \and Radu Timofte \and Luc Van Gool}
\institute{Computer Vision Lab, D-ITET, ETH Zurich, Switzerland\\stegeran@ethz.ch, \{radu.timofte, vangool\}@vision.ee.ethz.ch}

\maketitle

\begin{abstract}
Most face applications depend heavily on the accuracy of the face and facial landmarks detectors employed. Prediction of attributes such as gender, age, and identity usually completely fail when the faces are badly aligned due to inaccurate facial landmark detection.
Despite the impressive recent advances in face and facial landmark detection, little study is on the recovery from and detection of failures or inaccurate predictions.
In this work we study two top recent facial landmark detectors and devise confidence models for their outputs. We validate our failure detection approaches on standard benchmarks (AFLW, HELEN) and correctly identify more than 40\% of the failures in the outputs of the landmark detectors. Moreover, with our failure detection we can achieve a 12\% error reduction on a gender estimation application at the cost of a small increase in computation.
\end{abstract}

\section{Introduction}
\label{sec:introduction}
Face allows a non-invasive assessment of the human identity and personality. By looking to a face one can determine a large number of attributes such as identity of a person, biometrics (such as age, ethnicity, gender), facial expression, and various other facial features such as wearing lipstick, eyeglasses, piercings, etc. Therefore, in computer vision the automatic detection of faces from images and the estimation of such attributes are core tasks.

The past decades have shown tremendous advances in the field of object detection. More than a decade ago, Viola and Jones~\cite{Viola-CVPR-2001} were among the first to real-time accurately detect faces using cascaded detectors. Subsequent research led to large performance improvements for detection of a broad range of object classes~\cite{Felzenszwalb-PAMI-2010,Girshick-CVPR-2014}, including the articulated and challenging classes such as pedestrian~\cite{Benenson-CVPR-2012}.
Nowadays, the solutions to vision tasks are expected to be both highly accurate and with low time complexity.

For face attribute prediction the common pipeline is to first detect the face in the image space, then to detect facial landmarks or align a face model, and finally model such landmarks to predict facial attributes.
Each step is critical. Inaccurate face localization can propagate to erroneous detection of facial landmarks as the landmark detectors assume the presence of a full face in the input image region. Also, the accuracy of the facial landmarks strongly affects the performance of the attribute predictor as it relies on features extracted from (aligned and normalized) facial landmark regions. 

In this paper we study the detection of failures for facial landmark detectors and propose solutions at individual facial landmark level and at the whole face level. Detection of failures allows for recovery by using more robust but computationally demanding face detectors and/or alignment methods such that to reduce the errors of the subsequent processing steps (such as attribute predictors).

Our main contributions are:
\begin{enumerate}
\item We are the first to study failure detection in facial landmark detectors, to the best of our knowledge;
\item We propose failure detection methods for individual and groups of facial landmarks;
\item We achieve significant improvements on a gender prediction application with recovery from facial landmark failures at only small increase in computational time.
\end{enumerate}

\subsection{Related work}
\label{ssc:related_work}
Despite the dramatic propagation of errors in the forward face processing pipeline very few works focus on failure detection such that to recover and improve the performance at the end of the pipeline.

Most literature on failure detection concerns video and image sequence processing usually in the context of visual tracking where inaccurate localizations in each frame can lead to drifting and loose of track. One common model for increasing the stability and diminishing the drift or error accumulation is the forward-backward two-way checkup that effectively applies the tracking method twice in both directions from one frame to another to get the stable and reliable track~\cite{Kalal-ICPR-2010,Timofte-CVIU-2016}.
On single image processing the failure detection approaches from visual tracking are generally not applicable as they employ temporal information. 

Usually, in face detection and facial landmark literature the reduction of failures is the direct result of trading off the time complexity / running time of the methods. More complex models and intensive computations might allow for more robust performance. However, the costs for such reductions within the original models can be prohibitive for practical applications. For example, one might use a set of specialized detectors (or components) deployed together instead of a generic detector, and while the performance potentially can be improved, the time and memory complexities varies with the cardinality of the set~\cite{Zhu-CVPR-2012,Mathias-ECCV-2014,uricar2015real}.

The remainder of the paper is structured as follows. Section~\ref{sec:experimental_setup} briefly describes the datasets and facial landmark detectors from the experimental setup of our study.
Section~\ref{sec:failure_detection} studies the detection of failures for landmark detectors and introduces our methods. Section~\ref{sec:application} validates our failure detection on gender estimation, a typical facial attribute prediction application.
Section~\ref{sec:conclusion} concludes the paper.

\section{Experimental setup}
\label{sec:experimental_setup}
Before proceeding with our study we first introduce the experimental setup, that is the two fast and effective recent facial landmark detectors we work with (Uricar~\cite{uricar2015real} and Kazemi~\cite{kazemi2014one}) and the two of the most used recent datasets of face images with annotated facial landmarks (AFLW~\cite{Kostinger-ICCVW-2011} and HELEN~\cite{Le-ECCV-2012}).

\subsection{Facial landmark detectors}
\label{ssc:facial_landmark_detectors}

\begin{figure}
\centering
\includegraphics[width=0.7\textwidth]{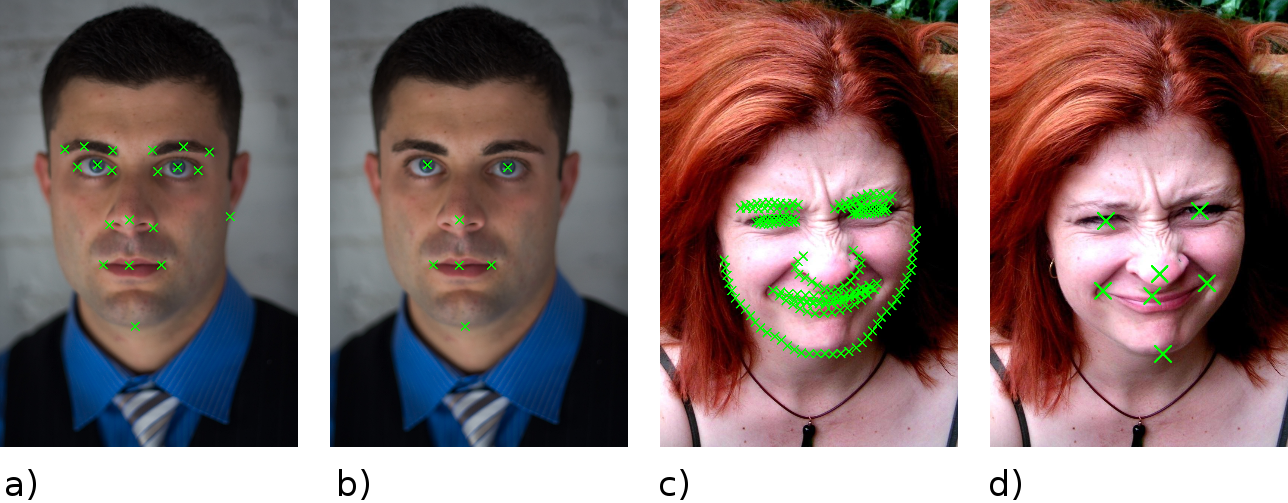}
\caption{
The images a) and c) show examples for the original annotations from AFLW~\cite{Kostinger-ICCVW-2011} and HELEN~\cite{Le-ECCV-2012}. Their normalized form which was used for all further computations is depicted in b) and d).
}
\label{fig:aflw_helen}
\end{figure}

\begin{figure}
\centering
\includegraphics[width=\textwidth]{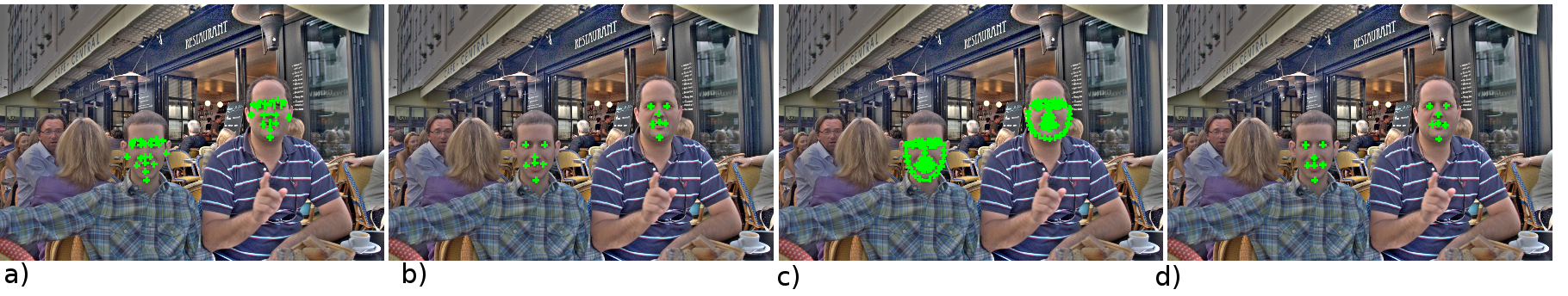}
\caption{
Landmark detections on an AFLW~\cite{Kostinger-ICCVW-2011} image. The original output from Uricar~\cite{uricar2015real} and Kazemi Detector~\cite{kazemi2014one} are shown in a) and c). Their normalized form which was used for all further computations is depicted in b) and d).
}
\label{fig:kazemi_uricar}
\end{figure}

\begin{figure}
\centering
\includegraphics[width=\textwidth]{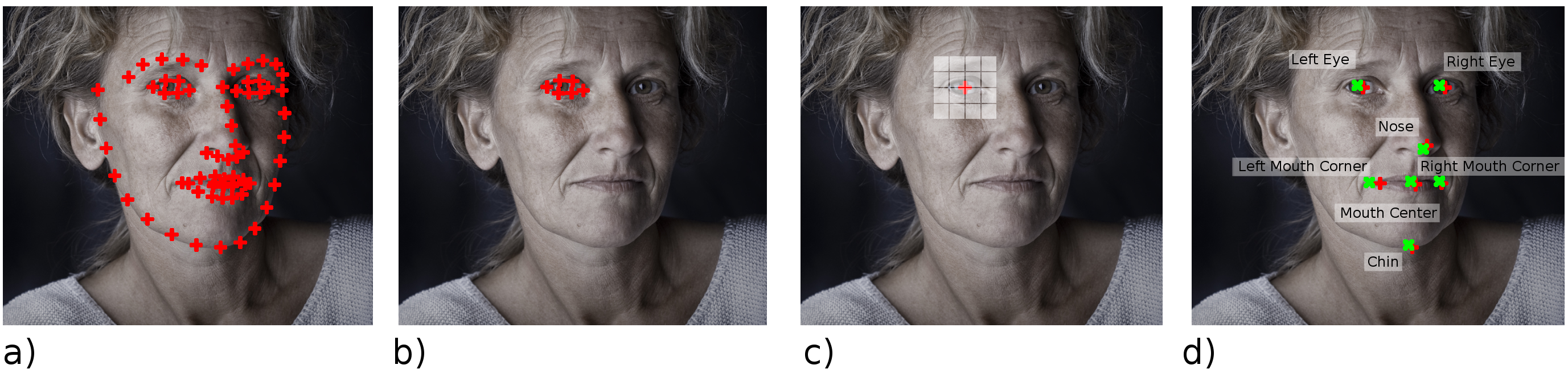}
\caption{
Image a) shows the landmarks predicted by Kazemi Detector~\cite{kazemi2014one}, b) shows an example for a landmark group, c) shows the average of the group of landmarks from the left eye and the corresponding grid for features extraction, d) shows the predicted landmarks marked with \textcolor{red}{red +} and the ground truth landmarks marked with \textcolor{green}{green x} and their names.
}
\label{fig:landmark_names}
\end{figure}

In order to validate our findings we use two fast and effective landmark detectors that are publicly available: 

\textbf{Uricar Detector} The real-time multi-view facial landmark detector learned by the Structured Output SVM of Uricar~\etal~\cite{uricar2015real} is the main facial landmark detector employed in our study. It uses the SO-SVM~\cite{Tsochantaridis-JMLR-2005} to learn the parameters of a Deformable Part Model~\cite{Felzenszwalb-PAMI-2010}. We directly use the authors' implementation ``uricamic/clandmark'' from github~\footnote{\url{https://github.com/uricamic/clandmark}} and the joint learned models for the front view which provides us 21 landmarks as shown in Figure~\ref{fig:kazemi_uricar}. In order to have a comparable result to Kazemi and Sullivan \cite{kazemi2014one} we also applied the dlib face detector. After using the same transformation method as shown in Figure~\ref{fig:landmark_names} we obtain the normalized landmarks as depicted in Figure~\ref{fig:kazemi_uricar} (b).

\textbf{Kazemi Detector} To validate the generality of our study and derived failure detection methods we also use the highly efficient Ensemble of Regression Trees (ERT) algorithm of Kazemi and Sullivan~\cite{kazemi2014one}. ERT can run in a millisecond for accurate face alignment results. It uses gradient boosting for learning an ensemble of regression trees~\cite{criminisi2010regression}. As in~\cite{kazemi2014one} we use the HELEN dataset for our experiments and the authors' dlib implementation~\footnote{\url{http://www.csc.kth.se/~vahidk/face_ert.html}} that provides 68 landmarks. The detected landmarks are depicted in Figure~\ref{fig:landmark_names} (a). We group them as shown for the left eye in (b) and then take the mean of each group as coordinate for that landmark (c). The picture in (d) shows the names of the used landmarks. We neglected the ears, because there are too few annotations (Section~\ref{ssc:datasets}).

\subsection{Datasets}
\label{ssc:datasets}
We use two popular facial landmark datasets: AFLW~\cite{Kostinger-ICCVW-2011} and HELEN~\cite{Le-ECCV-2012}. \\

\textbf{AFLW} dataset~\cite{Kostinger-ICCVW-2011} consists of 21k images with 24k annotated faces. It comes with a database that contains information about each face. The landmarks that we used from this database are: Left Eye Center, Right Eye Center, Nose Center, Mouth Left Corner, Mouth Right Corner, Mouth Center, Chin Center. From those we chose 5.5k faces that are aligned with pitch and yaw smaller than 15 degree. We used a test and validation set, each of 10\%, while the remaining 80\% was used for training. We did not use the ear landmarks, because about 40\% of them are missing. The miss rate for the chosen landmarks is below 3\%. Figure~\ref{fig:aflw_helen} shows an example for the original annotations of an AFLW image (a) and the annotations that we picked (b).

\textbf{HELEN} dataset~\cite{Le-ECCV-2012} consists of 2330 images which are splitted into a training set of 2000 images and a test set of 330 images. Moreover we split a validation set off the original training set with size of 10\% of the total number of images. Each image is annotated with 194 points. From those points we extracted the same landmarks as from AFLW. That extraction was done by choosing a group of annotations and taking the mean position like introduced in Section~\ref{ssc:facial_landmark_detectors}. An example for that is shown in Figure~\ref{fig:aflw_helen} (c) and (d). This method did not work well for the chin and the mouth center because annotation points with the same index are skewed for faces that are not in an aligned frontal view.

\subsection{Train, Test and Validation Set}
\label{testTrain}
When reporting results on data that were already used for training the model or choosing parameters, then it would be overoptimistic. Therefore, we split both datasets into training, validation and test set. Their sizes are listed in Section~\ref{ssc:datasets}. We trained the models for landmark confidence and gender prediction only on the training set. The model parameters were chosen in a 5 fold cross validation implemented in scikit-learn~\cite{scikit-learn}.
To choose the feature parameters, combination of features and combination of landmarks, we use the reported TrueCorrect95 on the validation set. The threshold for the predicted Confidence C is tuned on a disjunct set on the validation set. The final result in the application section is then reported on the test set.

\subsection{Measures}
\label{ssc:measures}

\begin{figure}
\centering
\includegraphics[width=0.4\textwidth]{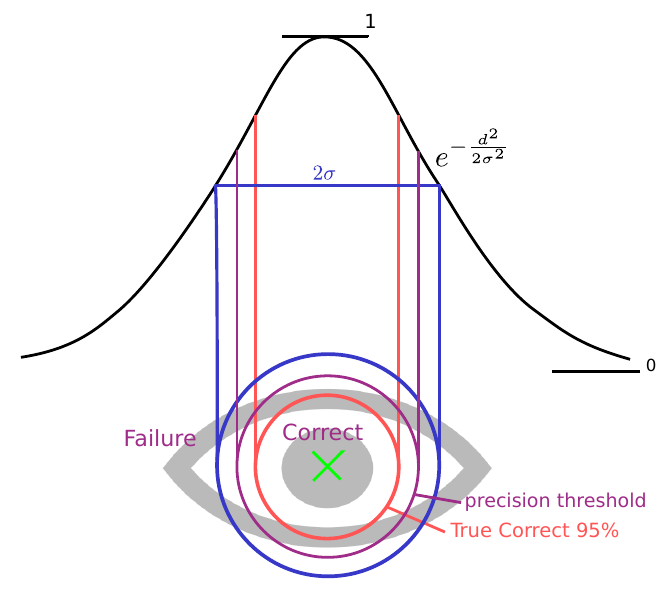}
\caption{
Dependency of Confidence (C) from the distance of ground truth (green X). The standard deviation $\sigma$ determines the width of the fitting function (blue), ground truth precision threshold (purple) and TrueCorrect95 (red).
}
\label{fig:thresholds}
\end{figure}

\textbf{Mean absolute error (MAE)} To measure the (pixel) accuracy of $n$ predicted landmark positions $\hat{Y}$ we calculate the standard mean absolute error (MAE) between the predictions and the known ground truth positions $Y$:
\begin{equation}
MAE=\frac{1}{n}\sum_{i=1}^{n}\sqrt{(\hat{Y}_{x,i}-Y_{x,i})^2 + (\hat{Y}_{y,i}-Y_{y,i})^2}
\end{equation}
The errors are generally small but there are some outliers and MAE is very sensitive with them. The following measure addresses this sensitivity. \\

\textbf{Confidence (C)} We need a transformation of the error in a way that small deviations of the predicted landmark from the ground truth have the highest dynamics. For this we use the Gaussian as depicted in Figure~\ref{fig:thresholds}, where $C \in (0,1]$ and $\sigma$ is the standard deviation:
\begin{equation}
C = e^{-\frac{(\hat{Y}-Y)^2}{2 \sigma ^2}}
\end{equation}

\textbf{Threshold (T)} Considering C measure for a fixed standard deviation $\sigma$ we can threshold it to obtain the plots from our experiments. 
A zero error level, corresponding to perfectly aligned landmarks, get the highest confidence $C=1$, while the failures get C values closer to 0.
The ground truth precision threshold in Figure~\ref{fig:thresholds} (purple) classifies the ground truth landmark confidence into correct and failure.

\textbf{Coefficient of determination ($R^2$)} For training all models we used the Coefficient of determination also called $R^2$.
$$ R^2 (y, \hat{y}) = 1 - \frac{\sum_{i=1}^{n}(y_i - \hat{y}_i)^2}{\sum_{i=1}^{n} (y_i - \overline{y})^2} $$

\textbf{TrueCorrect95} Rate of failed landmarks that are detected as a failure at an operating point where the rate of correct landmarks that are detected as correct is above 95\%. It determines the threshold for the predicted confidence that is depicted in Figure~\ref{fig:thresholds} (red). It classifies the detected landmarks into detected as correct and detected as failure. When reporting the TrueCorrect95 we split 20\% off the set on which we report to tune the threshold for the predicted Confidence C.

\textbf{Runtime} All runtimes were measured on an Intel Core i7-5500U @ 2.4 GHz processor when using 2 threads for calculation.

\section{Failure detection}
\label{sec:failure_detection}

\subsection{Facial features}
\label{ssc:features}

Before extracting the features the face is rotated and cropped according to the predicted eyes annotations. We use a patch size of 128x128 pixels for the face plus a border of 48 pixels in each direction to be able to apply descriptors also near the edge. In order to avoid the trade off between descriptiveness and speed that CNN features have, we used handcrafted features instead. Therefore we used a combination of SIFT, HoG and LBP in order to get very descriptive features that can be efficiently calculated on CPUs. \\

\textbf{PCA} For all descriptors listed below that have a feature vector larger than 1500, we apply a PCA that is learned from the training data to reduce the runtime and memory usage of the predictor training. To do so we use the scikit-learn~\cite{scikit-learn} PCA implementation that reduces the dimensionality to 1500.

\textbf{HoG} Navneet Dalal and Bill Triggs~\cite{dalal2005histograms} showed that the combination of a grid of histogram of oriented gradients descriptors and an SVM can outperform other feature sets for human detection. Inspired by that we use the implementation of HoG form scikit-image to apply the same principle to a patch around the landmarks. We tuned the parameters patch size $\in \{\frac{1}{8}, \frac{2}{8}, \frac{3}{8}, \frac{4}{8} \}$ of the face, number of cells in the patch $\in \{1, 2^2, 4^2, 8^2\}$ and number of orientations $\in \{4, 8\}$. The best result for each Landmark can be seen in Table \ref{tab:tune_features_AFLW} and Table \ref{tab:tune_features_HELEN}.

\textbf{LBP} Local Binary Pattern calculates the number of neighbor pixels that have a greater gray scale value than the center. We used the same setup as for HoG with a grid of cells for which we calculated the LBP for all pixels. For each of those cells we calculated the histogram and used that as features. We tuned the parameters patch size $\in \{\frac{1}{8}, \frac{2}{8}, \frac{3}{8}, \frac{4}{8} \}$, number of cells in the patch $\in \{1, 2^2, 4^2, 8^2\}$ and the radius $\in \{1, 2, 3, 4\}$ in which the LBP is calculated. For that we used the scikit-image library function ``local\_binary\_pattern'' with the method parameter ``uniform''~\cite{scikit-learn}. The best result for each Landmark can be seen in Table \ref{tab:tune_features_AFLW} and Table \ref{tab:tune_features_HELEN}.

\textbf{SIFT} The Scale-invariant feature transform is based on a grid of HoG. It is largely invariant to illumination and local affine distortions~\cite{lowe1999object}. For SIFT we used the OpenCV implementation and applied the same method with the grid as for HoG and LBP. We tuned the parameters patch size $\in \{\frac{1}{8}, \frac{2}{8}, \frac{3}{8}, \frac{4}{8} \}$ and number of cells in the patch $\in \{1, 2, 4, 8\}$. The best result for each Landmark can be seen in Table \ref{tab:tune_features_AFLW} and Table \ref{tab:tune_features_HELEN}.

\begin{table}[]
\centering
\caption{Validation results on AFLW for SIFT, HoG and LBP using different parameters. `O' stands for Orientations and `TC95' for TrueCorrect95.}
\label{tab:tune_features_AFLW}
\begin{tabular}{l||rcrrr|rcrr|cccrr}
         & \multicolumn{5}{c}{SIFT} &   \multicolumn{4}{|c}{HoG} &\multicolumn{5}{|c}{LBP}\\          Landmark &  Size &  Cells &  $R^2$ &  TC95  &  O &  Size &  Cells &  $R^2$ &  TC95 &  Radius &  Size &  Cells &  $R^2$ &  TC95 \\
\midrule                                                                                                         
   chinC &   2.5 &      $4^2$ & -0.74 &          0.40  &             4 &   3.8 &      $8^2$ & -1.95 &        0.35 &       3 &   3.8 &      $8^2$ & -6.6 &         0.30 \\
    eyeL &   1.2 &      $4^2$ &  0.37 &          0.52  &             8 &   2.5 &      $4^2$ &  0.06 &        0.48 &       2 &   2.5 &      $8^2$ & -2.4 & \textbf{0.45} \\
    eyeR &   1.2 &      $4^2$ &  0.37 & \textbf{0.60}  &             4 &   3.8 &      $8^2$ &  0.12 &        0.51 &       3 &   2.5 &      $8^2$ & -2.1 &         0.43 \\
  mouthC &   1.2 &      $4^2$ & -0.25 &          0.57  &             8 &   3.8 &      $4^2$ & -0.14 &\textbf{0.56}&       3 &   2.5 &      $8^2$ & -4.8 &         0.34 \\
  mouthL &   3.8 &      $4^2$ &  0.21 &          0.50  &             4 &   5.0 &      $8^2$ & -0.24 &        0.46 &       4 &   3.8 &      $2^2$ & -2.4 &         0.32 \\
  mouthR &   2.5 &      $4^2$ &  0.13 &          0.42  &             8 &   1.2 &      $2^2$ & -1.62 &        0.43 &       4 &   2.5 &      $8^2$ & -3.5 &         0.41 \\
   noseC &   1.2 &      $2^2$ &  0.20 &          0.55  &             8 &   2.5 &      $1^2$ & -3.06 &        0.38 &       3 &   2.5 &      $4^2$ & -2.0 &         0.34 \\
\end{tabular}
  \vspace{-0.5cm}
\end{table}

\begin{table}[]
\centering
\caption{Validation results on HELEN for SIFT, HoG and LBP using different parameters. `O' stands for Orientations and `TC95' for TrueCorrect95.}
\label{tab:tune_features_HELEN}
\resizebox{\linewidth}{!}{
\begin{tabular}{l||rcrrr|rcrr|cccrr}
         & \multicolumn{5}{c}{SIFT} &   \multicolumn{4}{|c}{HoG} &\multicolumn{5}{|c}{LBP}\\          
Landmark &  Size &  Cells &  $R^2$ &  TC95 &  O &  Size &  Cells &  $R^2$ &  TC95 &  Radius &  Size &  Cells &  $R^2$ &  TC95 \\
\midrule                                                                                                        
   chinC &   2.5 &      $4^2$ & -1.08 &        0.37 &             4 &   5.0 &      $8^2$ & -0.809 &         0.29 &       3 &   2.5 &      $8^2$ & -4.24 &         0.27 \\
    eyeL &   3.8 &      $4^2$ &  0.50 &        0.59 &             4 &   5.0 &      $4^2$ &  0.457 &         0.70 &       2 &   2.5 &      $4^2$ & -0.54 &         0.40 \\
    eyeR &   1.2 &      $4^2$ &  0.66 &        0.68 &             4 &   5.0 &      $8^2$ &  0.589 &\textbf{0.77} &       4 &   3.8 &      $4^2$ & -0.83 &\textbf{0.57} \\
  mouthC &   2.5 &      $4^2$ &  0.20 &        0.55 &             8 &   5.0 &      $8^2$ &  0.051 &         0.50 &       4 &   5.0 &      $8^2$ & -1.86 &         0.31 \\
  mouthL &   1.2 &      $4^2$ &  0.39 &\textbf{0.77}&             4 &   3.8 &      $8^2$ &  0.024 &         0.60 &       4 &   2.5 &      $8^2$ & -1.89 &         0.27 \\
  mouthR &   3.8 &      $4^2$ & -0.10 &        0.51 &             8 &   3.8 &      $4^2$ & -0.389 &         0.47 &       3 &   2.5 &      $8^2$ & -1.63 &         0.27 \\
   noseC &   2.5 &      $4^2$ &  0.67 &        0.74 &             8 &   5.0 &      $4^2$ &  0.519 &         0.75 &       3 &   5.0 &      $4^2$ & -0.70 &         0.44 \\
\end{tabular}
}
  \vspace{-0.5cm}
\end{table}

\subsection{Individual Confidence of landmark Detectors}
\label{ssc:confidence_landmark}

In this section we describe how we predict the $R^2$ score that is explained in Section~\ref{ssc:measures} by using the features from Section~\ref{ssc:features} for a single landmark at a time. \\

As training data we used the annotations from AFLW~\cite{Kostinger-ICCVW-2011} and HELEN~\cite{Le-ECCV-2012} as described in Section~\ref{ssc:datasets}. In order to train a robust predictor we generate 5 different annotations per face from the data set. We introduced a Gaussian error in distance to the actual position with a standard deviation of 10\% of the face size. That corresponds to 13 pixels of the 128 pixel face patch side length. The angle was chosen uniformly randomly. \\

We use the Support Vector Regression (SVR) as predictor for the confidence and therefore have to tune the C and $\epsilon$ parameter, as well as the kernel type. To do so we used a 5 fold cross validation. The Parameters that we used in the exhaustive search are $C \in \{0.3, 0.5, 0.7\}$, $\epsilon \in \{0.01, 0.05, 0.1\}$, kernel $\in $\{Radial Basis Function, Linear\}. We chose the scikit-learn~\cite{scikit-learn} implementation for the exhaustive search, cross validation and the SVR. The SVR is based on libsvm~\cite{LIBSVM}. \\

\textbf{Feature Parameters Search}
First we extracted all feature descriptors for each parameter configuration and each landmark as described in Section~\ref{ssc:features}. Each of those features were then used to train the SVR in a exhaustive search for the best parameters. The results which are reported on the validation set can be seen in Table \ref{tab:tune_features_AFLW} and \ref{tab:tune_features_HELEN}. According to the TrueCorrect95 scores for the AFLW dataset SIFT is the best or almost the best feature descriptor for all landmarks. For HELEN this is the case for five of seven landmarks. The size of the training and validation set was 2300 and 480 respectively. \\

\textbf{Feature Combination Search}
Secondly we searched over all combinations of different descriptors for each landmark. For both datasets we used about 2300 samples for training. For each sample the feature extraction for the best combination of features and all landmarks takes 107ms (Section \ref{ssc:measures}). The result which is reported on the validation set can be seen in Table \ref{tab:feat_combo}. It shows that in 86\% of the cases SIFT is contained in the best feature combination. \\

Several parameters influence the failure detection accuracy measurement.
First the standard deviation that is used to fit the confidence C (Section~\ref{ssc:measures}) for both the ground truth MAE and the predicted MAE. This was set to 10\% of the face size.
Secondly the ground truth precision threshold, which determines the ground truth correct and failures. It was set to 0.65 at which both the predicted landmarks form the Uricar~\cite{uricar2015real} and Kazemi~\cite{kazemi2014one} Detector have an error rate that is not too unbalanced. That confidence C corresponds to a distance error of about 9.2\%. Figure~\ref{fig:tune_precision_lm} reports the resulting error rate on a range from 0\% to 40\% of the face size.
Thirdly the prediction precision threshold has to be set to an operating point where the rate of correctly marked as correct is above 95\%. The tuning and reporting is done on disjunct subsets of the validation set. The tuning for the fixed ground truth precision threshold of 0.65 is shown in Figure~\ref{fig:tune_threshold_lm}. The operating point that is then used to calculate the TrueCorrect95 is marked in those plots.

\begin{table}[]
\centering
\caption{Results for best two Feature Descriptor combinations for each Landmark individually on the AFLW Dataset}
\label{tab:feat_combo}
\begin{tabular}{l||lrr|lrr}
  		 &    \multicolumn{3}{|c|}{AFLW}                &     \multicolumn{3}{c}{HELEN}\\
Landmark &     Descriptors &     $R^2$ &  TrueCorrect95 &     Descriptors &     $R^2$ &  TrueCorrect95 \\
\midrule                                                 
   chinC &       hog, sift & -0.54 &         0.35 &            sift & -0.60 &        0.26 \\
   chinC &            sift & -0.48 &         0.33 &             hog & -0.89 &        0.25 \\
    eyeL &            sift &  0.43 & \textbf{0.75}&        hog, lbp &  0.39 &        0.63 \\
    eyeL &       hog, sift &  0.44 &         0.73 &  hog, sift, lbp &  0.54 &        0.59 \\
    eyeR &       hog, sift &  0.38 &         0.62 &        hog, lbp &  0.54 &        0.75 \\
    eyeR &            sift &  0.36 &         0.55 &       hog, sift &  0.72 &        0.74 \\
  mouthC &            sift & -0.05 &         0.50 &       sift, lbp & -0.18 &        0.55 \\
  mouthC &       hog, sift & -0.01 &         0.49 &       hog, sift &  0.29 &        0.54 \\
  mouthL &  hog, sift, lbp &  0.27 &         0.61 &       hog, sift &  0.37 &        0.59 \\
  mouthL &            sift &  0.26 &         0.59 &            sift &  0.35 &        0.54 \\
  mouthR &            sift &  0.15 &         0.57 &            sift &  0.06 &        0.45 \\
  mouthR &       hog, sift &  0.16 &         0.56 &       hog, sift &  0.07 &        0.44 \\
   noseC &       sift, lbp &  0.27 &         0.59 &  hog, sift, lbp &  0.68 &\textbf{0.80} \\
   noseC &  hog, sift, lbp &  0.27 &         0.59 &       hog, sift &  0.68 &        0.79 \\
\end{tabular}
  \vspace{-0.5cm}
\end{table}

\subsection{Joint confidence of landmark Detectors}
\label{ssc:confidence_face}

\begin{figure}[th!]
\centering
\includegraphics[width=0.8\textwidth]{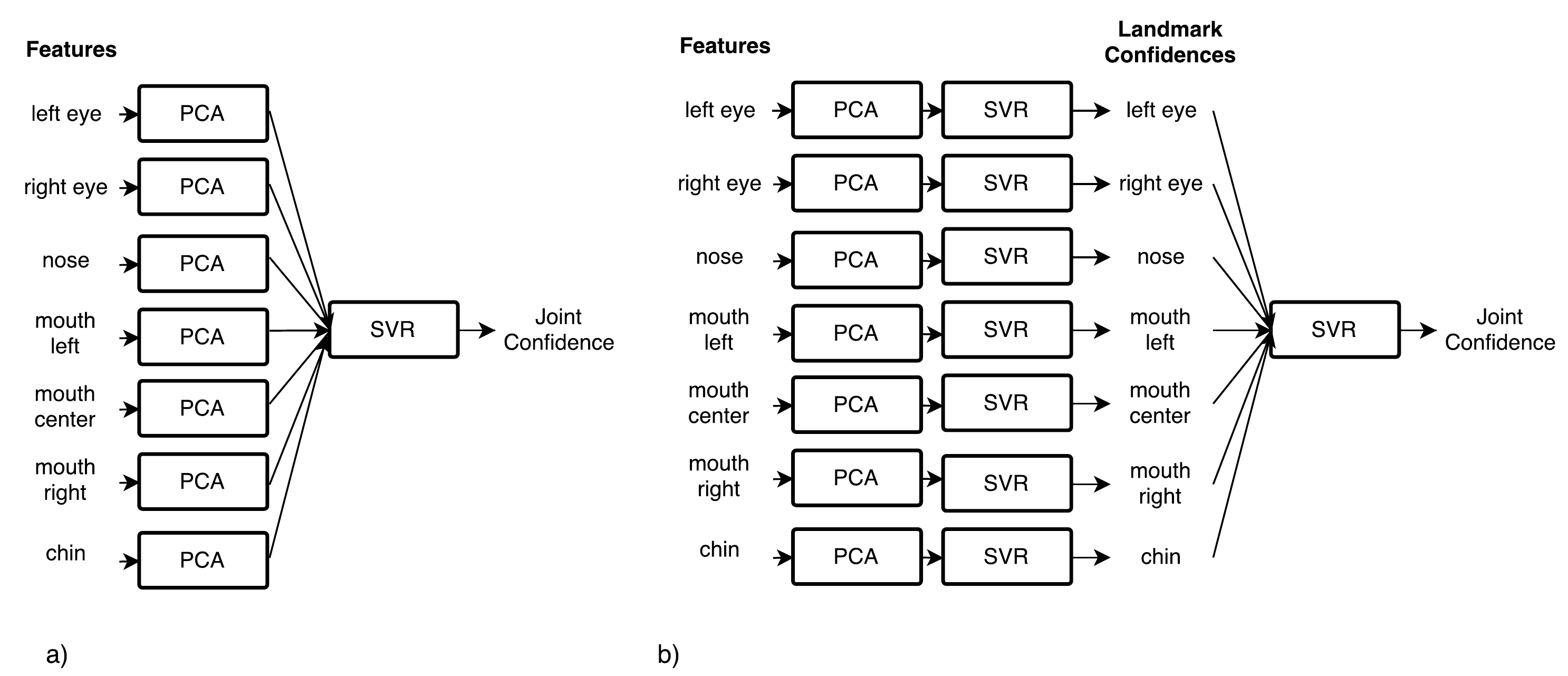}
\caption{
Figure a) shows the configuration where all features are jointly feeded into the SVR and b) the Cascaded SVR.
}
\label{fig:svm_config}
\vspace{-0.5cm}
\end{figure}

In order to increase the accuracy of failure detection, we used different combinations of facial landmarks to train the predictor. The two setups that we used are shown in Figure~\ref{fig:svm_config}. As training data we used again the annotations from AFLW~\cite{Kostinger-ICCVW-2011} and HELEN~\cite{Le-ECCV-2012} as described in Section~\ref{ssc:datasets} and generate landmarks similarly as in Section~\ref{ssc:confidence_landmark}. The difference is that we have to model a predicted face annotation that is both shifted overall and individually for each landmark compared to the ground truth. So we use the superposition of two errors. One is generated for each landmark with a standard deviation of 7\% of the face size. The other is generated once for the whole face with a standard deviation of 10\% of the face size. Both have a uniformly randomly generated angle. The total error is the superposition of both. The ground truth confidence C was calculated from the MAE of the generated landmarks with a precision of 10\% of the face size.\\

\begin{figure}[ht]
  \centering
  \begin{subfigure}[b]{0.33\linewidth}
    \centering\includegraphics[width=\linewidth]{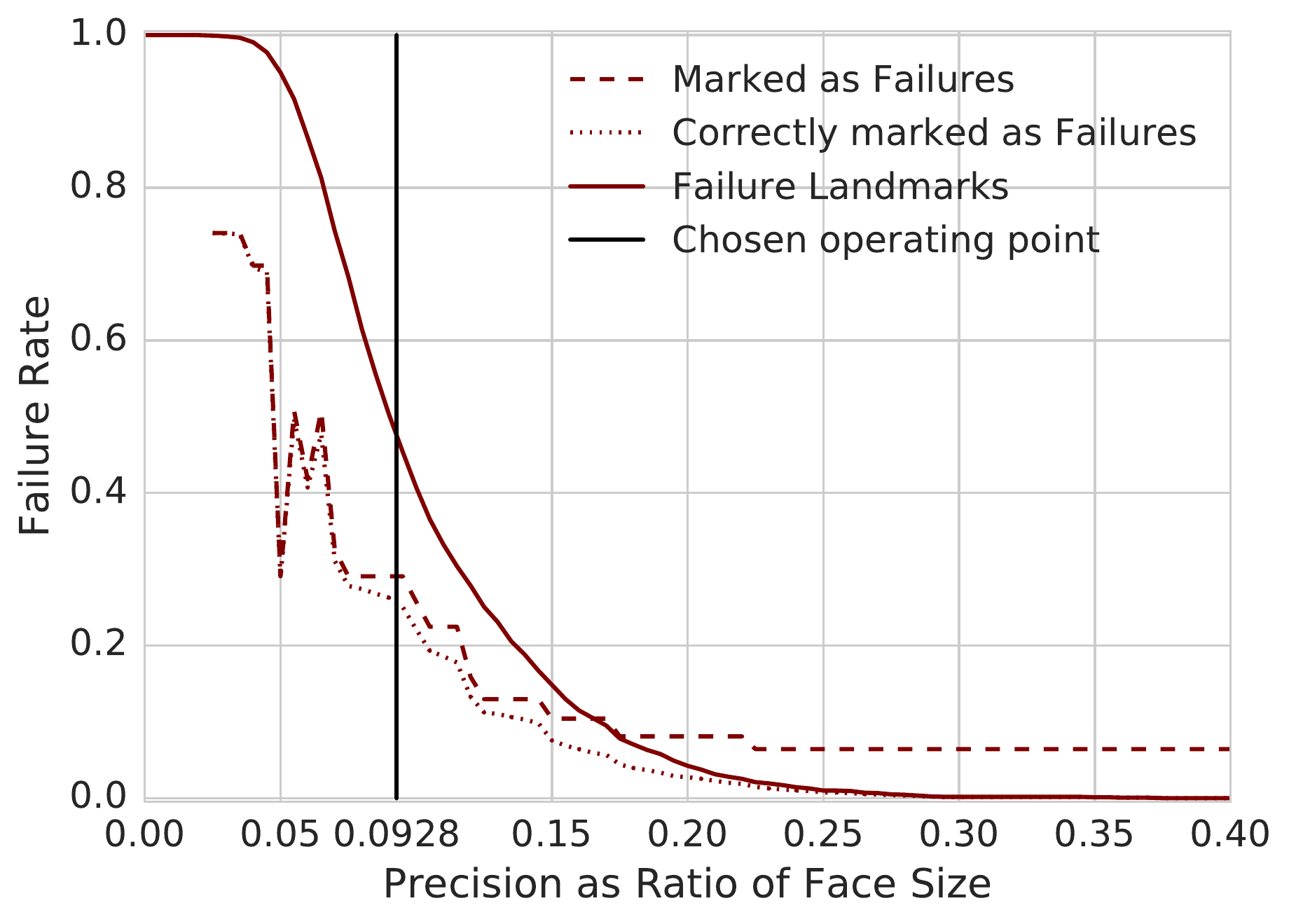}
    \caption{\label{fig:tune_precision_lm_gen} {\tiny Generated Landmarks on AFLW}}
  \end{subfigure}%
  \begin{subfigure}[b]{0.33\linewidth}
    \centering\includegraphics[width=\linewidth]{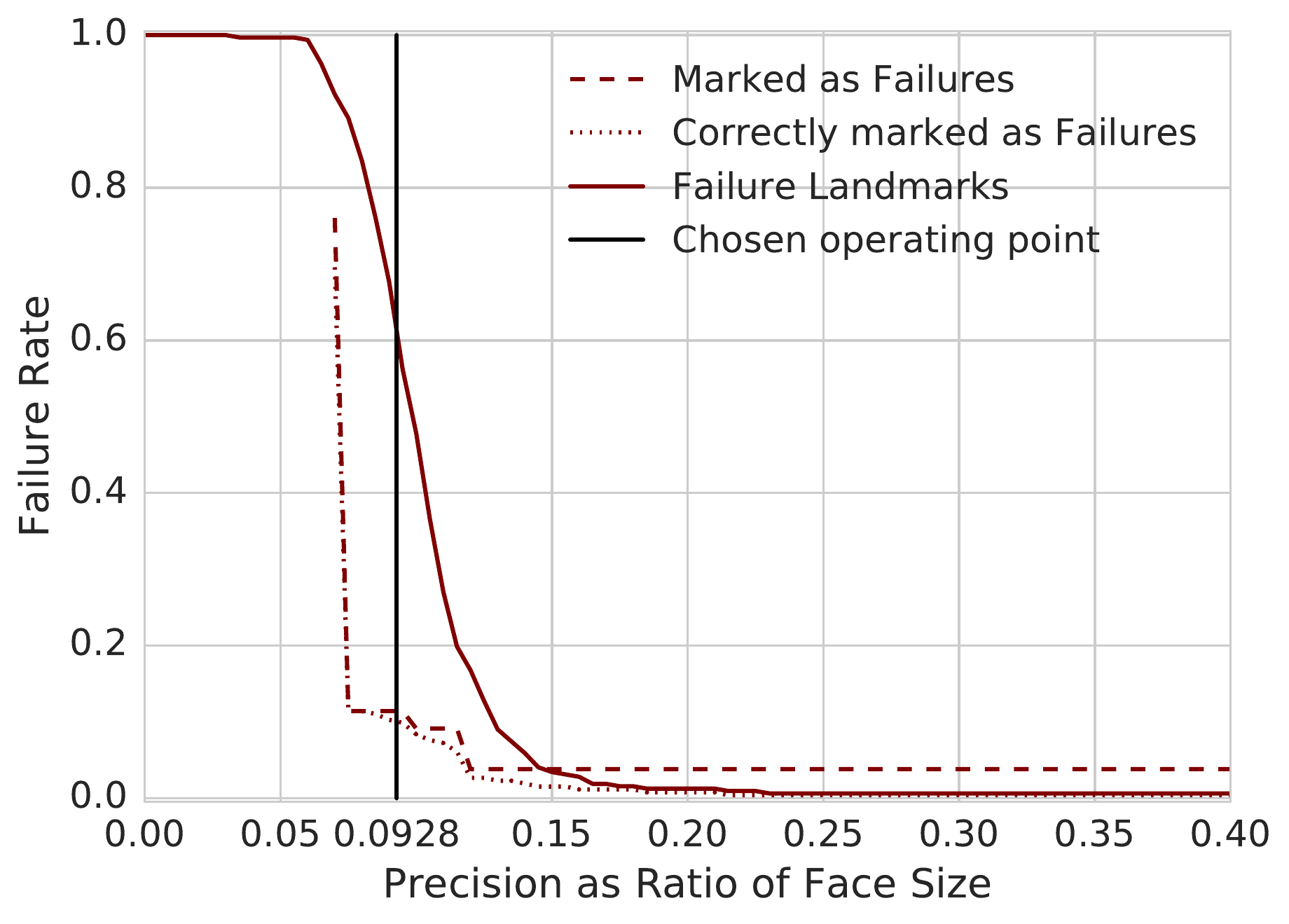}
    \caption{\label{fig:tune_precision_lm_uricar}{\tiny Uricar Detector on AFLW}}
  \end{subfigure}%
  \begin{subfigure}[b]{0.33\linewidth}
    \centering\includegraphics[width=\linewidth]{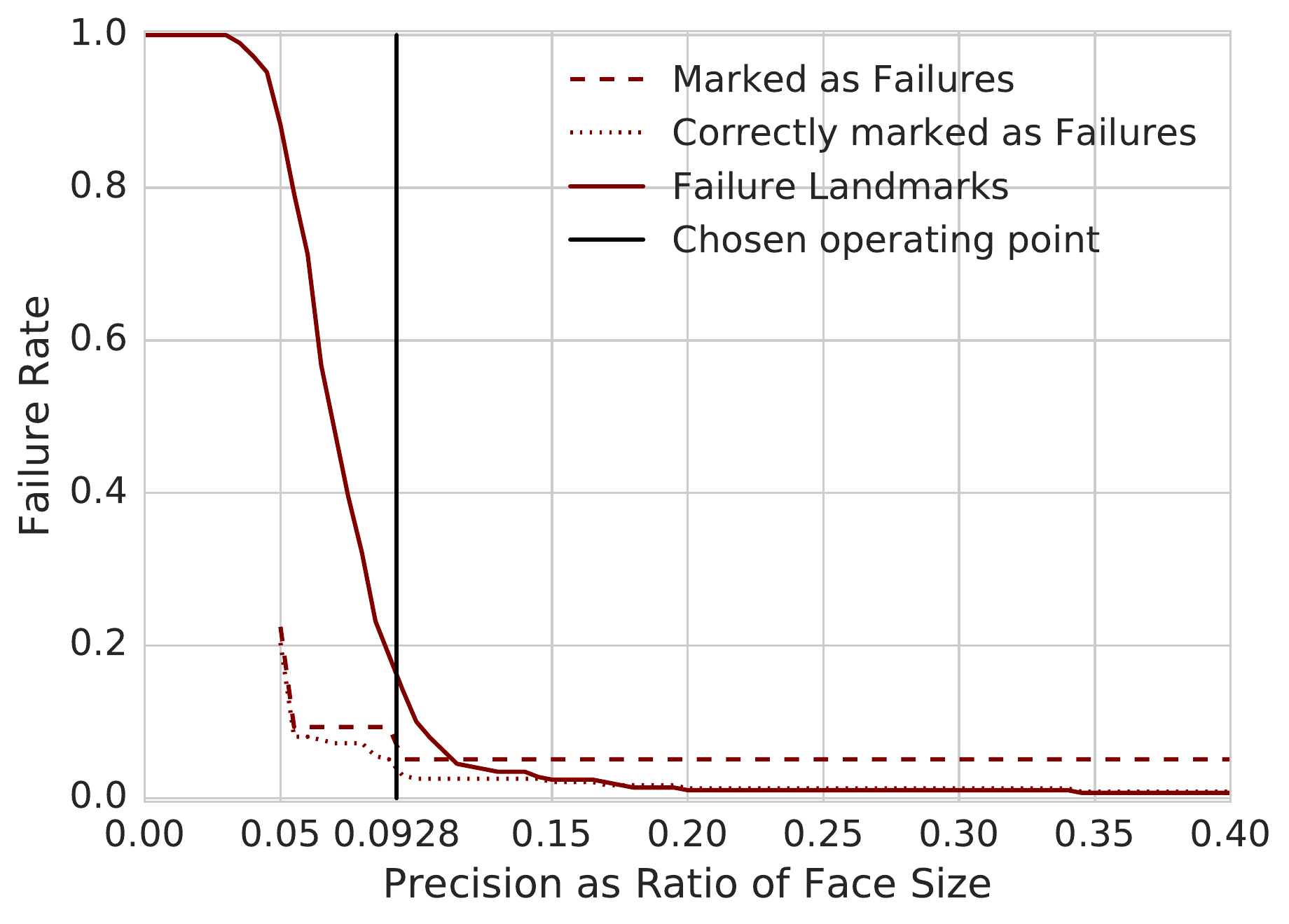}
    \caption{\label{fig:tune_precision_lm_kazemi}{\tiny Kazemi Detector on AFLW}}
  \end{subfigure}
  
  \begin{subfigure}[b]{0.33\linewidth}
    \centering\includegraphics[width=\linewidth]{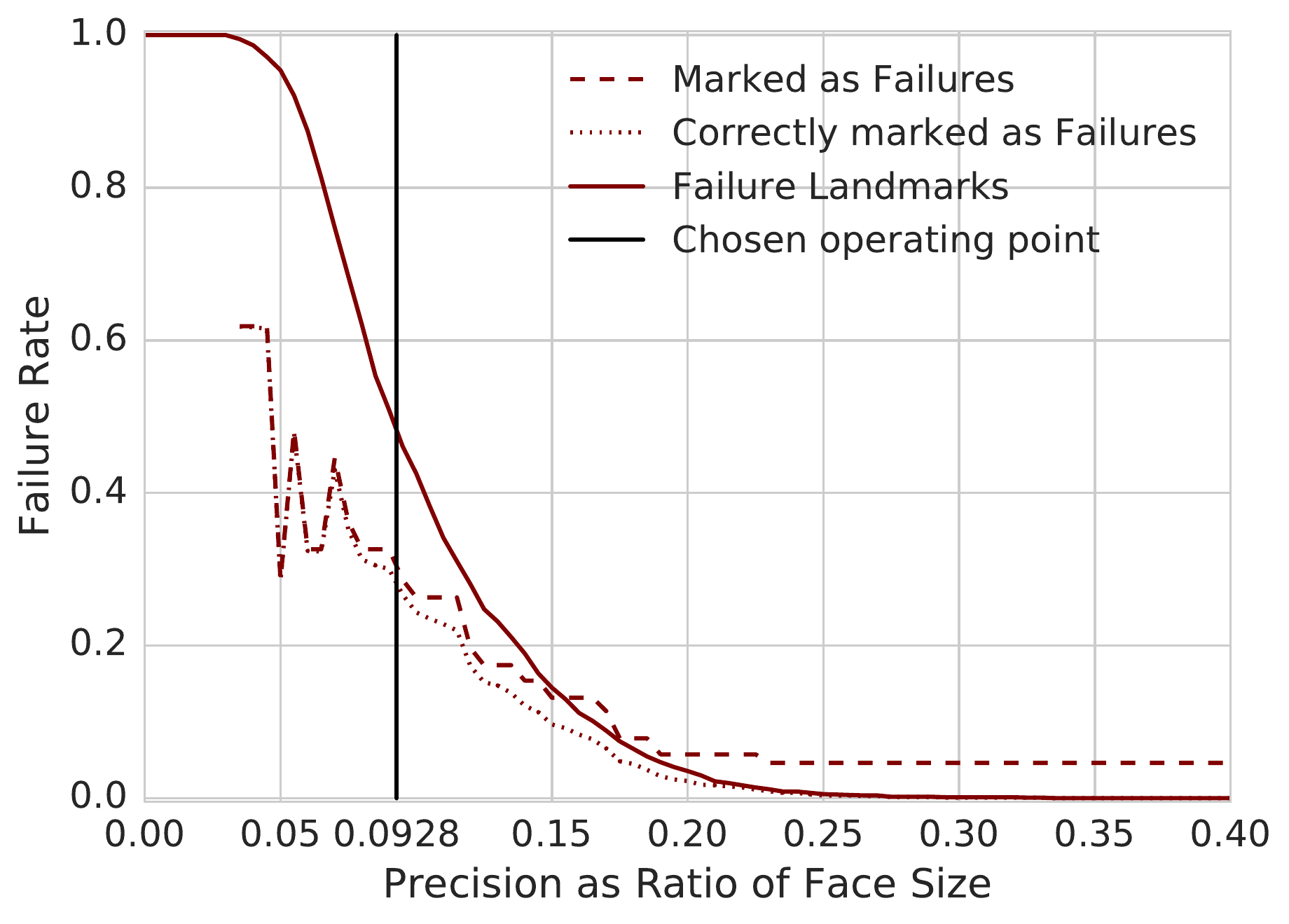}
    \caption{\label{fig:tune_precision_lm_gen_h}{\tiny Generated Landmarks on HELEN}}
  \end{subfigure}%
  \begin{subfigure}[b]{0.33\linewidth}
    \centering\includegraphics[width=\linewidth]{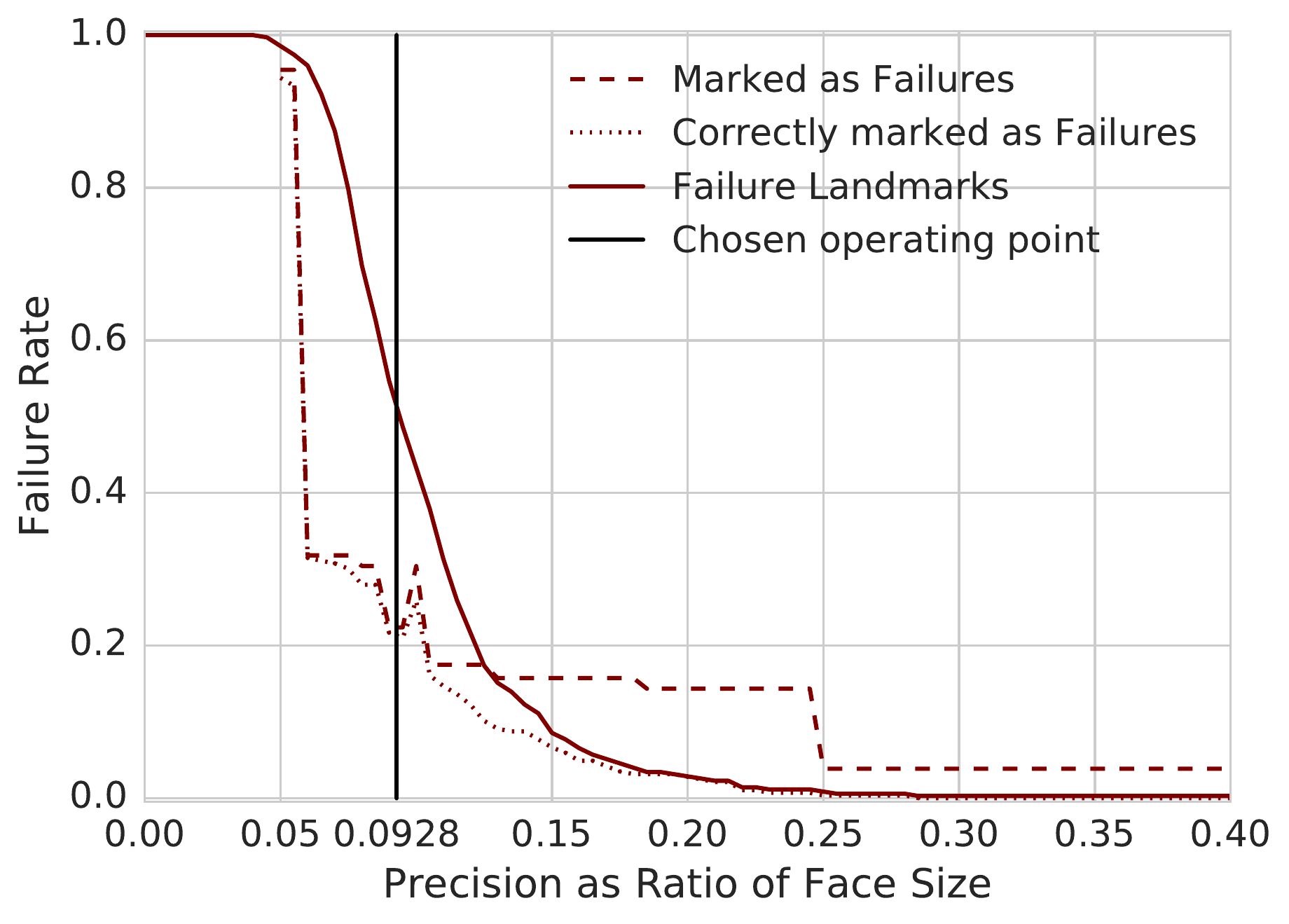}
    \caption{\label{fig:tune_precision_lm_uricar_h}{\tiny Uricar Detector on HELEN}}
  \end{subfigure}%
  \begin{subfigure}[b]{0.33\linewidth}
    \centering\includegraphics[width=\linewidth]{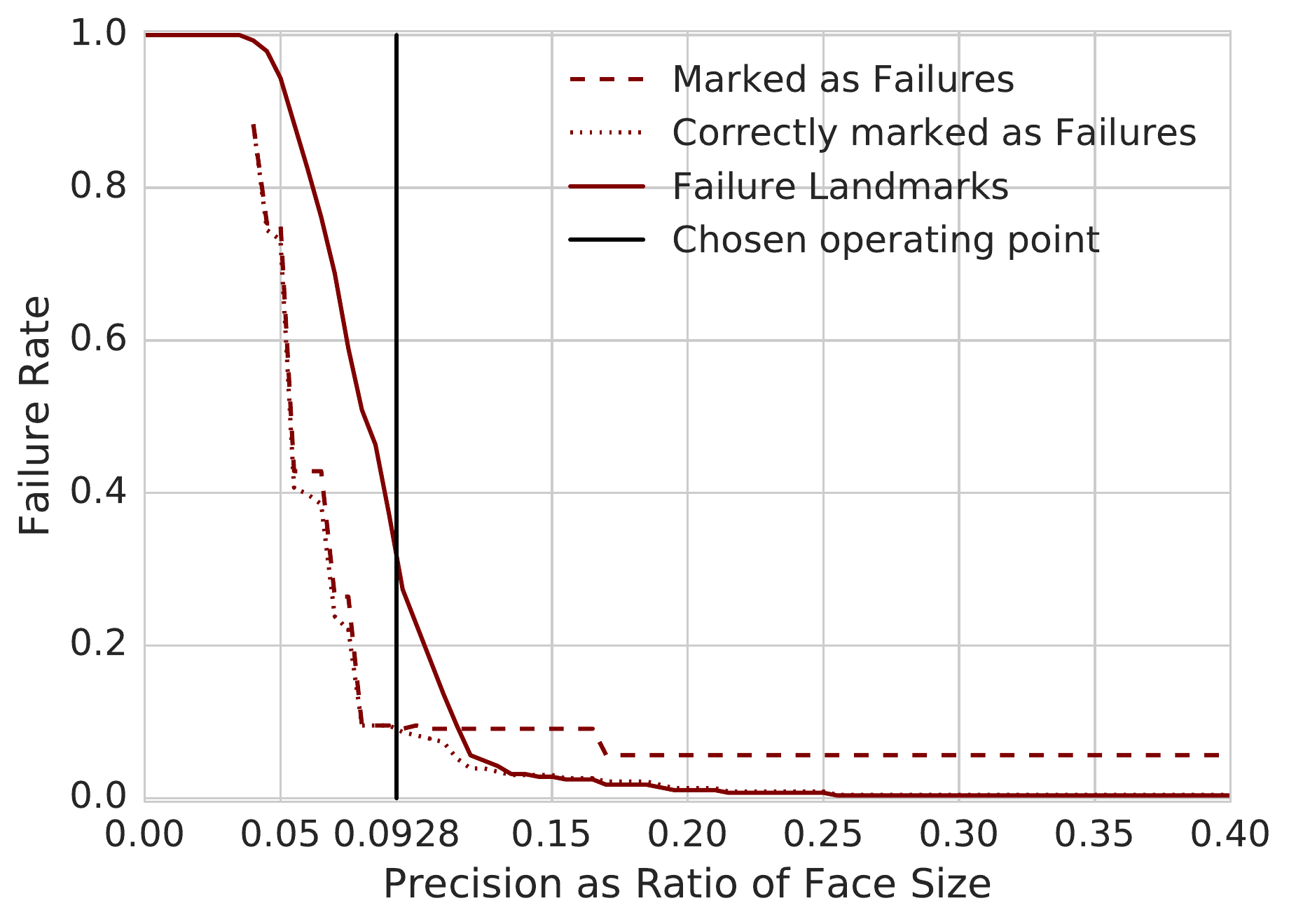}
    \caption{\label{fig:tune_precision_lm_kazemi_h}{\tiny Kazemi Detector on HELEN}}
  \end{subfigure}
  \caption{\label{fig:tune_precision_lm}
  The failure rate is the result of tuning the threshold to a value where the rate of correct landmarks that are marked as correct is above 95\%. All plots were generated with the model that is described in this section and show the combination of nose, left mouth corner and left eye.
  }
\vspace{-0.5cm}
\end{figure}

\begin{figure}[ht]
  \centering
  \begin{subfigure}[b]{0.33\linewidth}
    \centering\includegraphics[width=\linewidth]{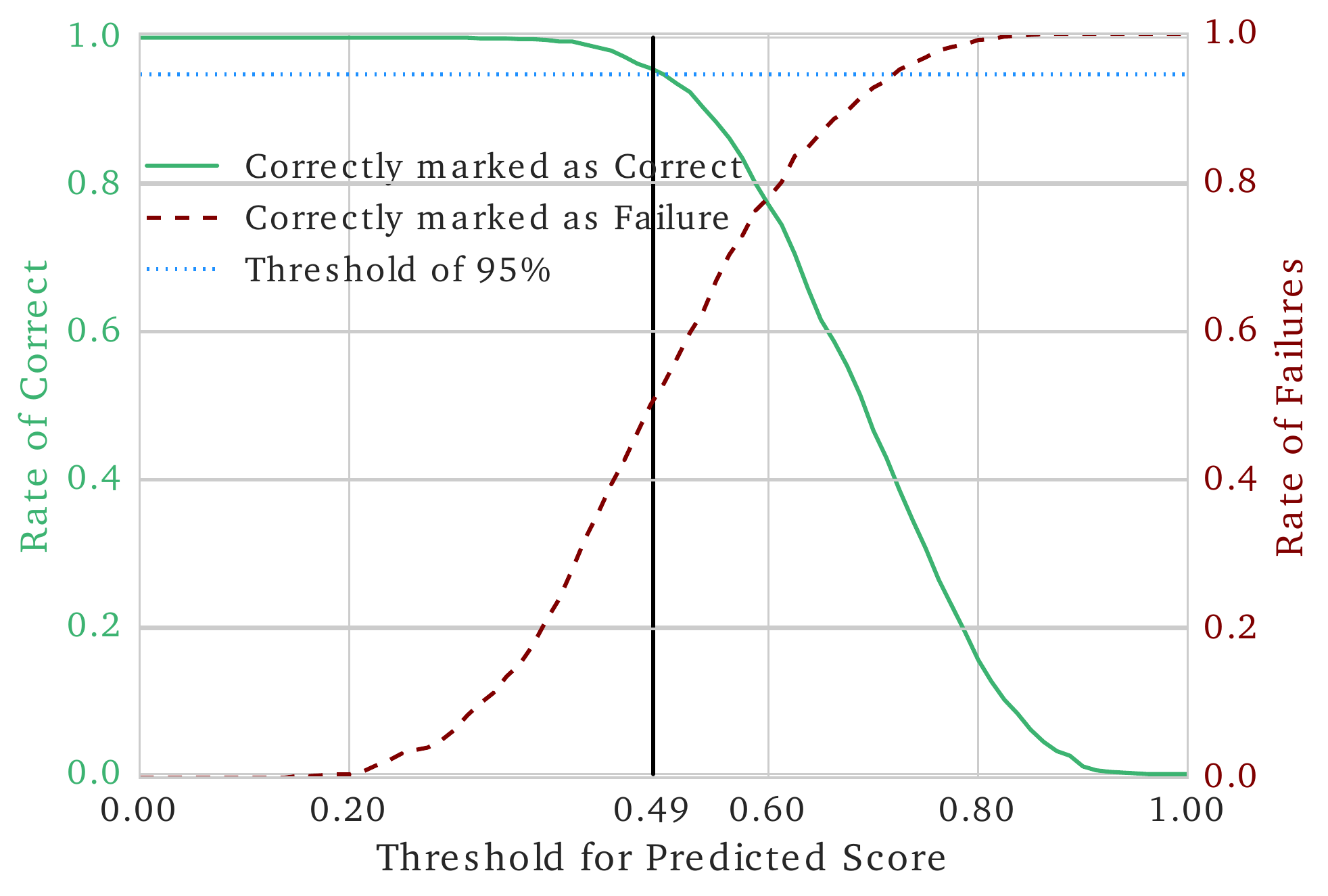}
    \caption{\label{fig:tune_threshold_lm_gen}{\tiny Generated Landmarks on AFLW}}
  \end{subfigure}%
  \begin{subfigure}[b]{0.33\linewidth}
    \centering\includegraphics[width=\linewidth]{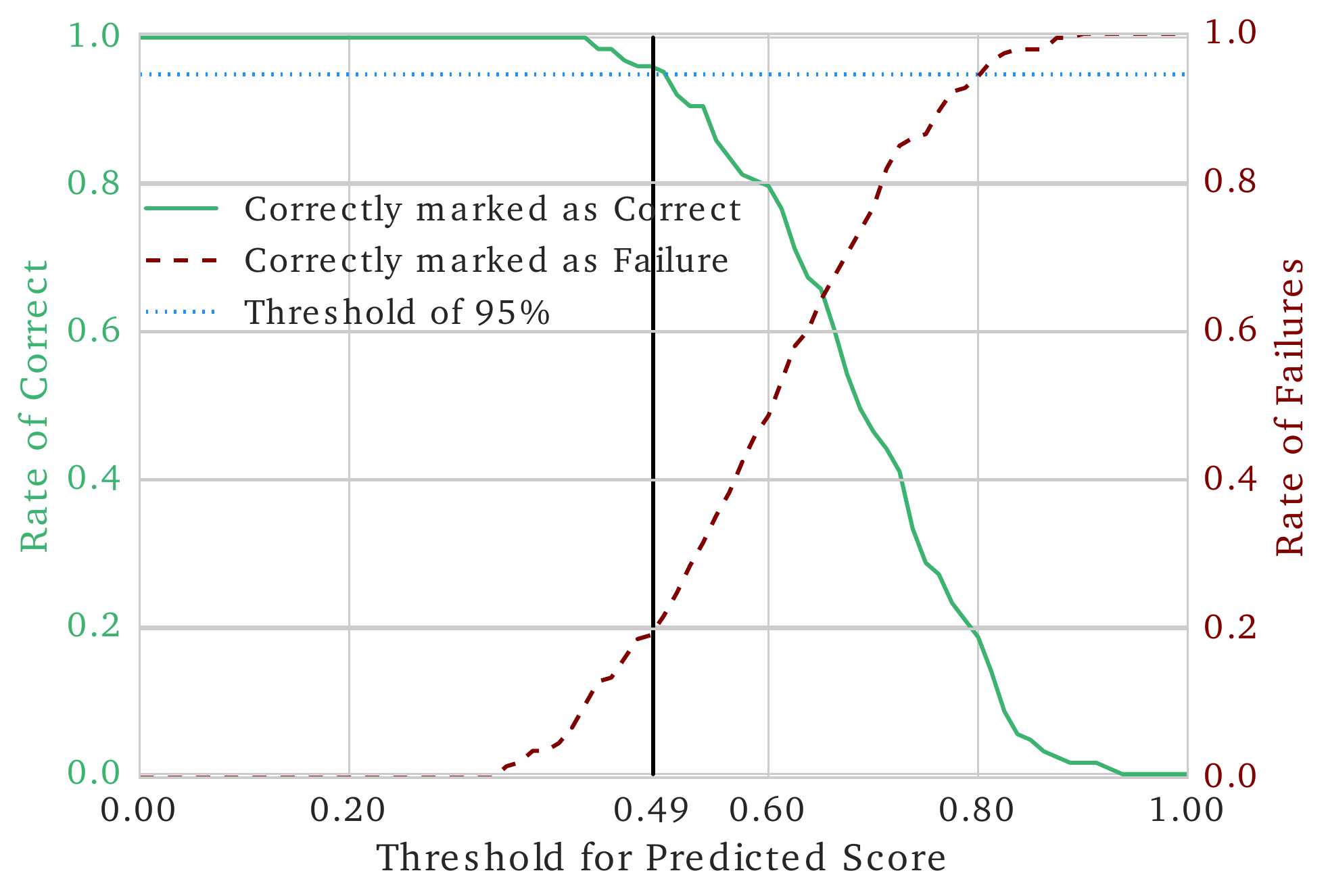}
    \caption{\label{fig:tune_threshold_lm_uricar}{\tiny Uricar Detector on AFLW}}
  \end{subfigure}%
  \begin{subfigure}[b]{0.33\linewidth}
    \centering\includegraphics[width=\linewidth]{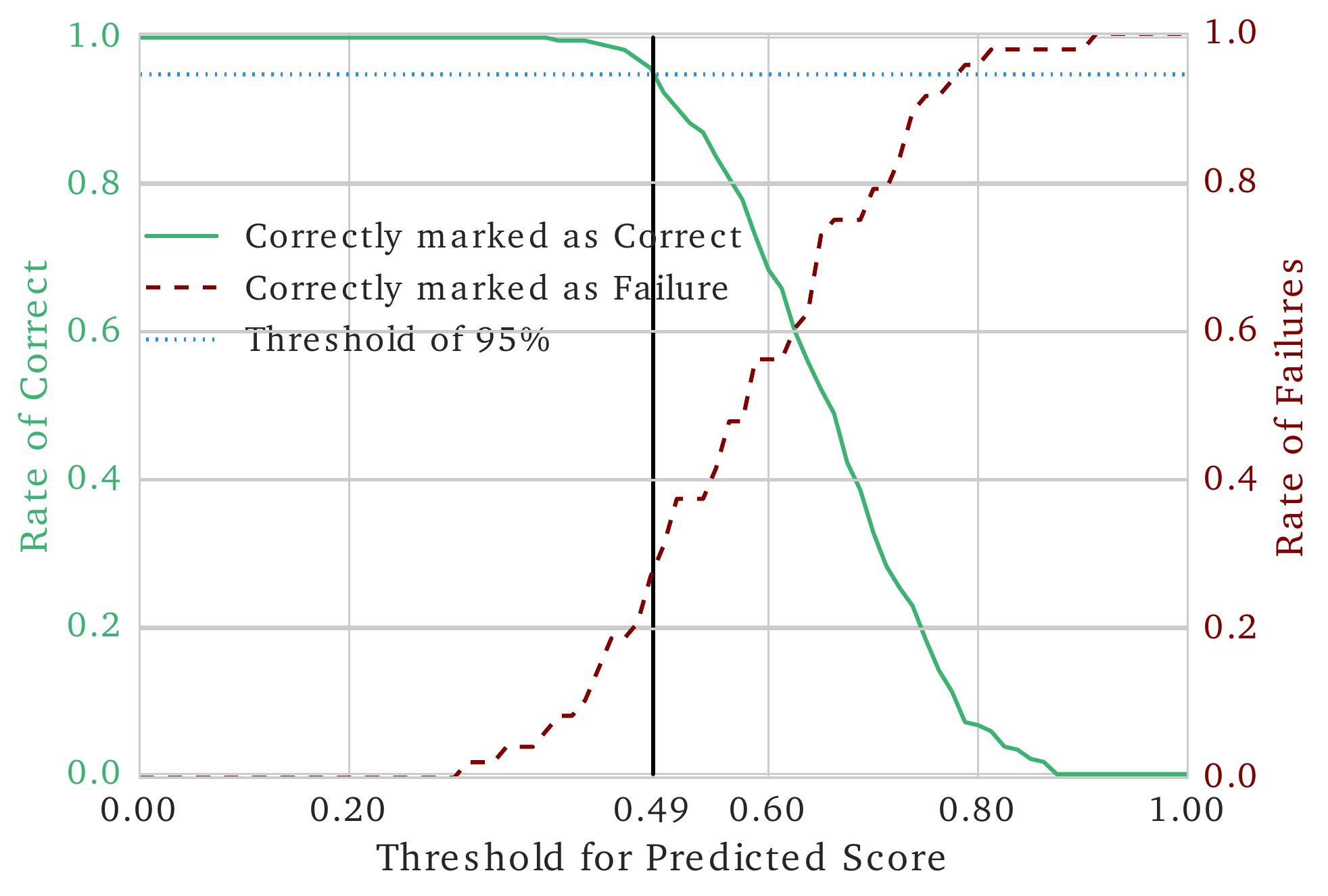}
    \caption{\label{fig:tune_threshold_lm_kazemi}{\tiny Kazemi Detector on AFLW}}
  \end{subfigure}
  
  \begin{subfigure}[b]{0.33\linewidth}
    \centering\includegraphics[width=\linewidth]{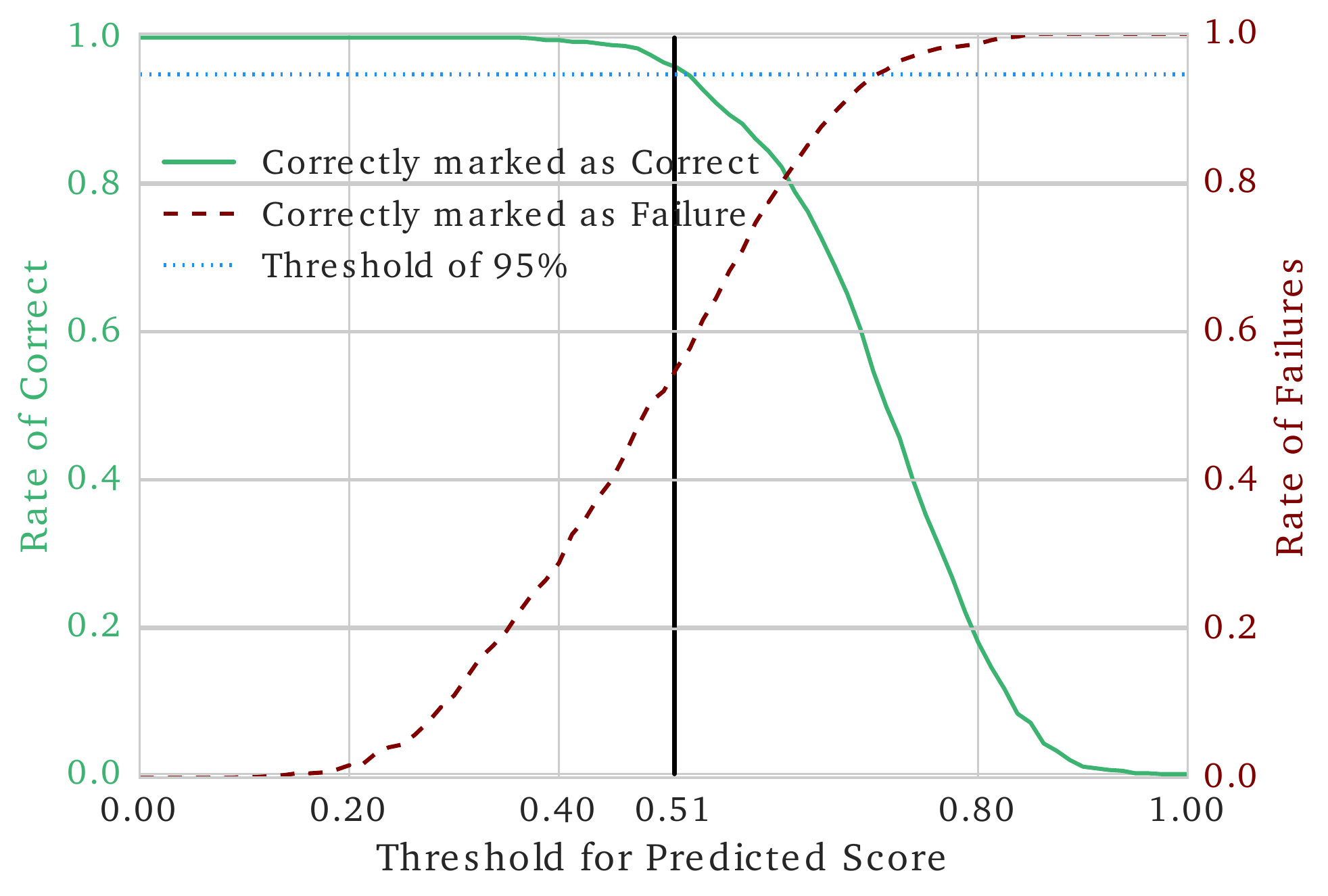}
    \caption{\label{fig:tune_threshold_lm_gen_h}{\tiny Generated Landmarks on HELEN}}
  \end{subfigure}%
  \begin{subfigure}[b]{0.33\linewidth}
    \centering\includegraphics[width=\linewidth]{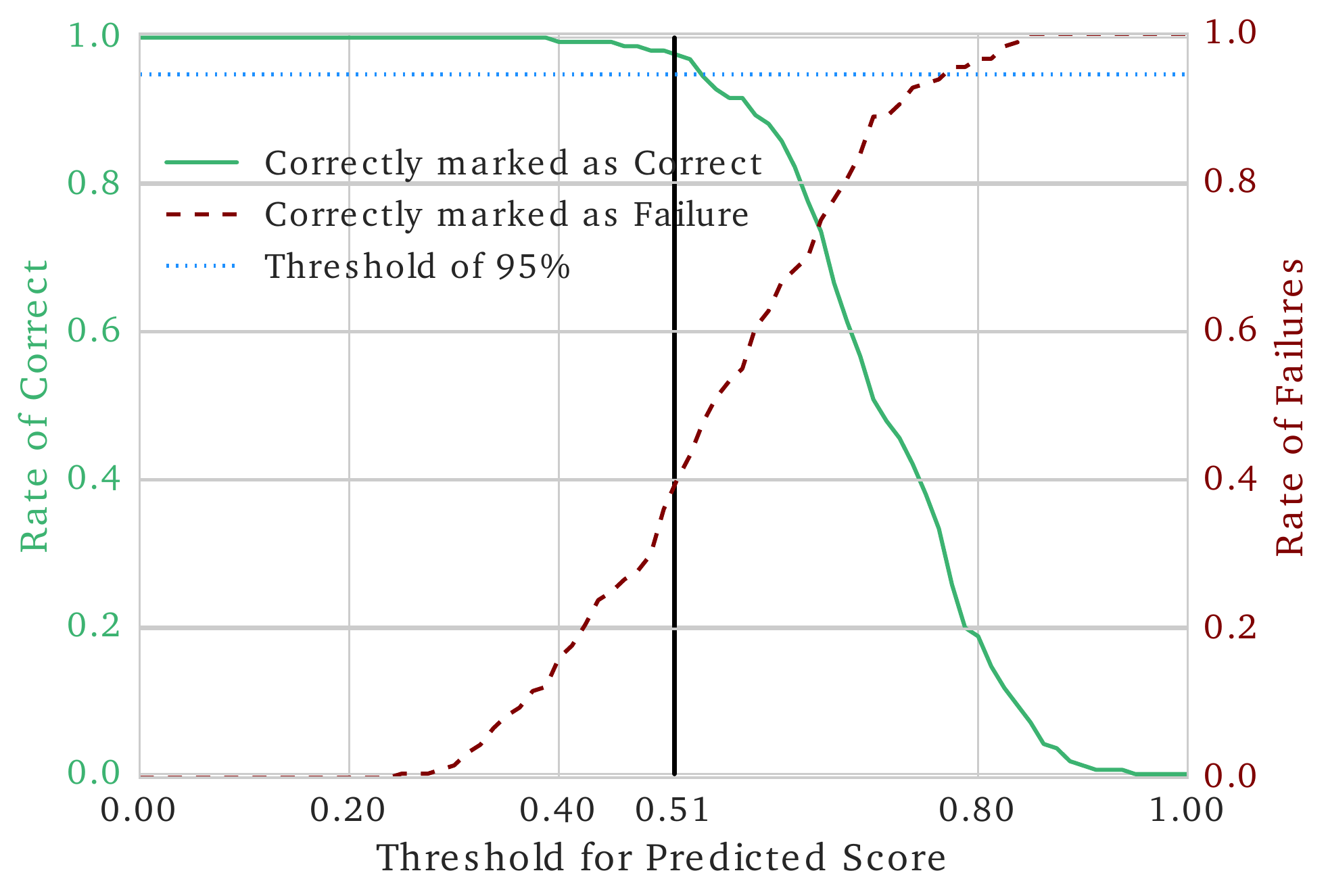}
    \caption{\label{fig:tune_threshold_lm_uricar_h}{\tiny Uricar Detector on HELEN}}
  \end{subfigure}%
  \begin{subfigure}[b]{0.33\linewidth}
    \centering\includegraphics[width=\linewidth]{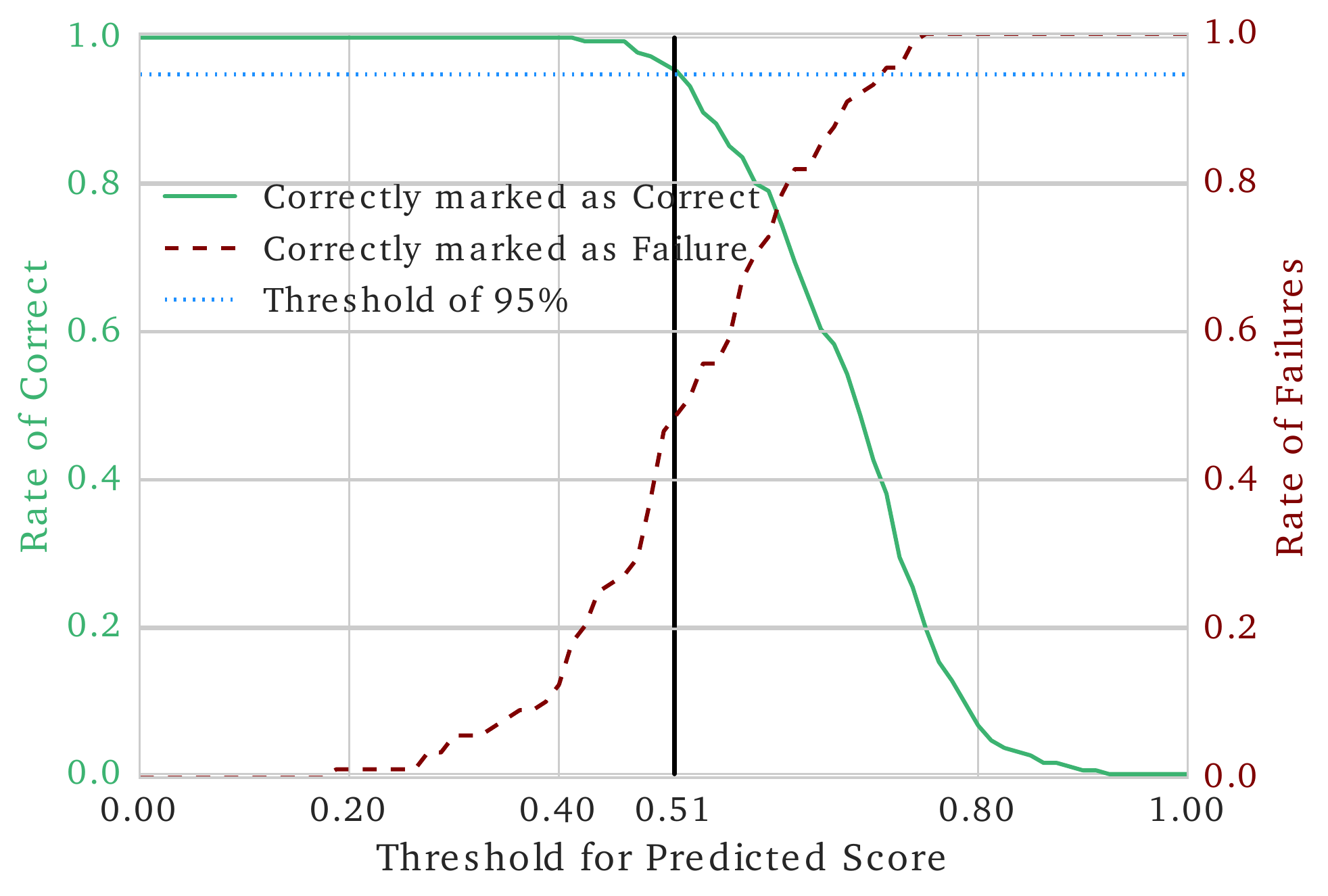}
    \caption{\label{fig:tune_threshold_lm_kazemi_h}{\tiny Kazemi Detector on HELEN}}
  \end{subfigure}
  
  \caption{\label{fig:tune_threshold_lm}
  The failure rate is shown at fixed ground truth precision threshold of 0.65 for different prediction thresholds. All plots were generated with the same model that is described in this section and show the combination of nose, left mouth corner and left eye.
  }
  \vspace{-0.5cm}
\end{figure}

\begin{figure}[ht]
  \centering
  \begin{subfigure}[b]{0.25\linewidth}
    \centering\includegraphics[width=\linewidth]{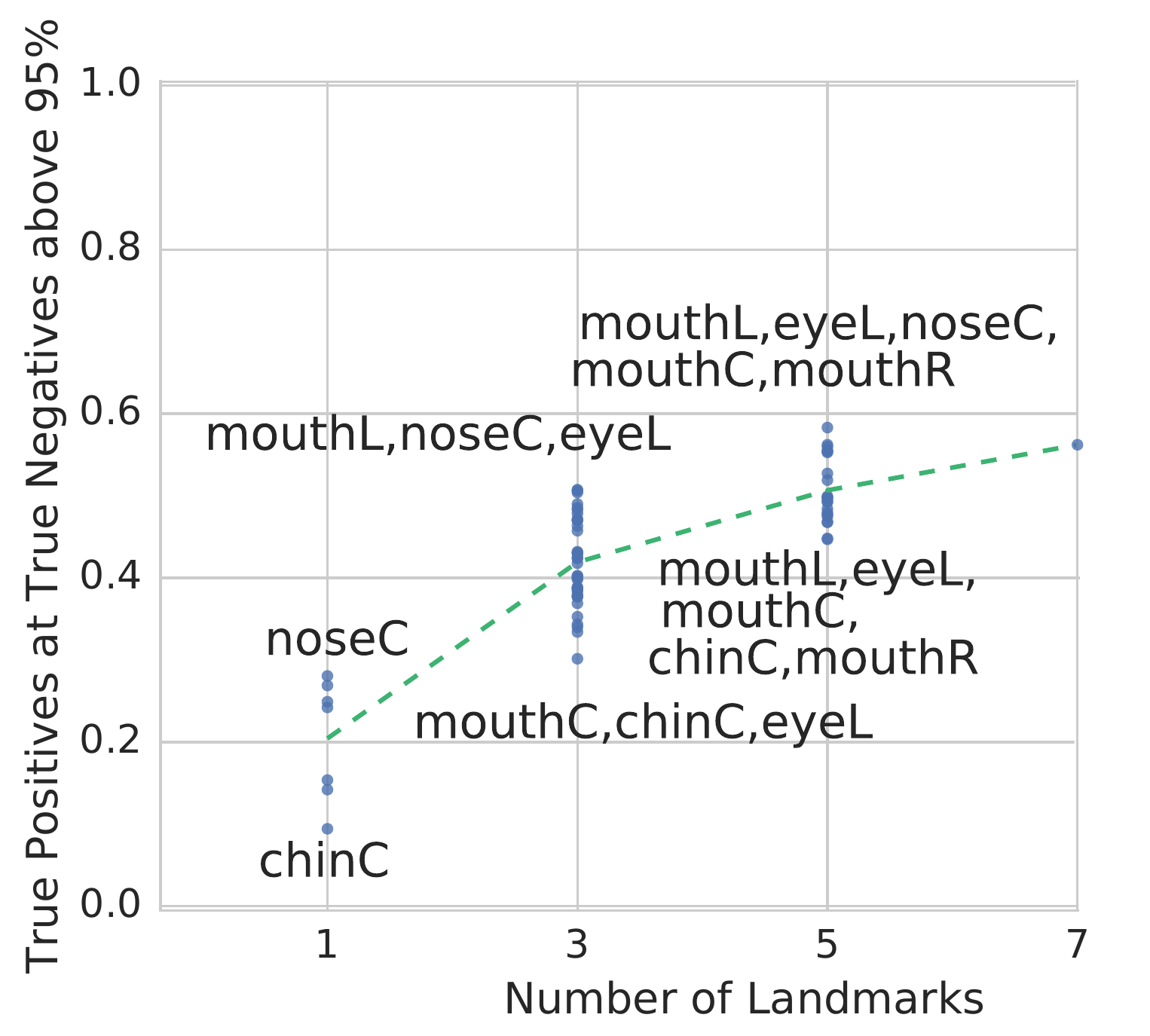}
    \caption{\label{fig:AFLW-gen-scatter_plot}{\tiny Gen. AFLW}}
  \end{subfigure}%
  \begin{subfigure}[b]{0.25\linewidth}
    \centering\includegraphics[width=\linewidth]{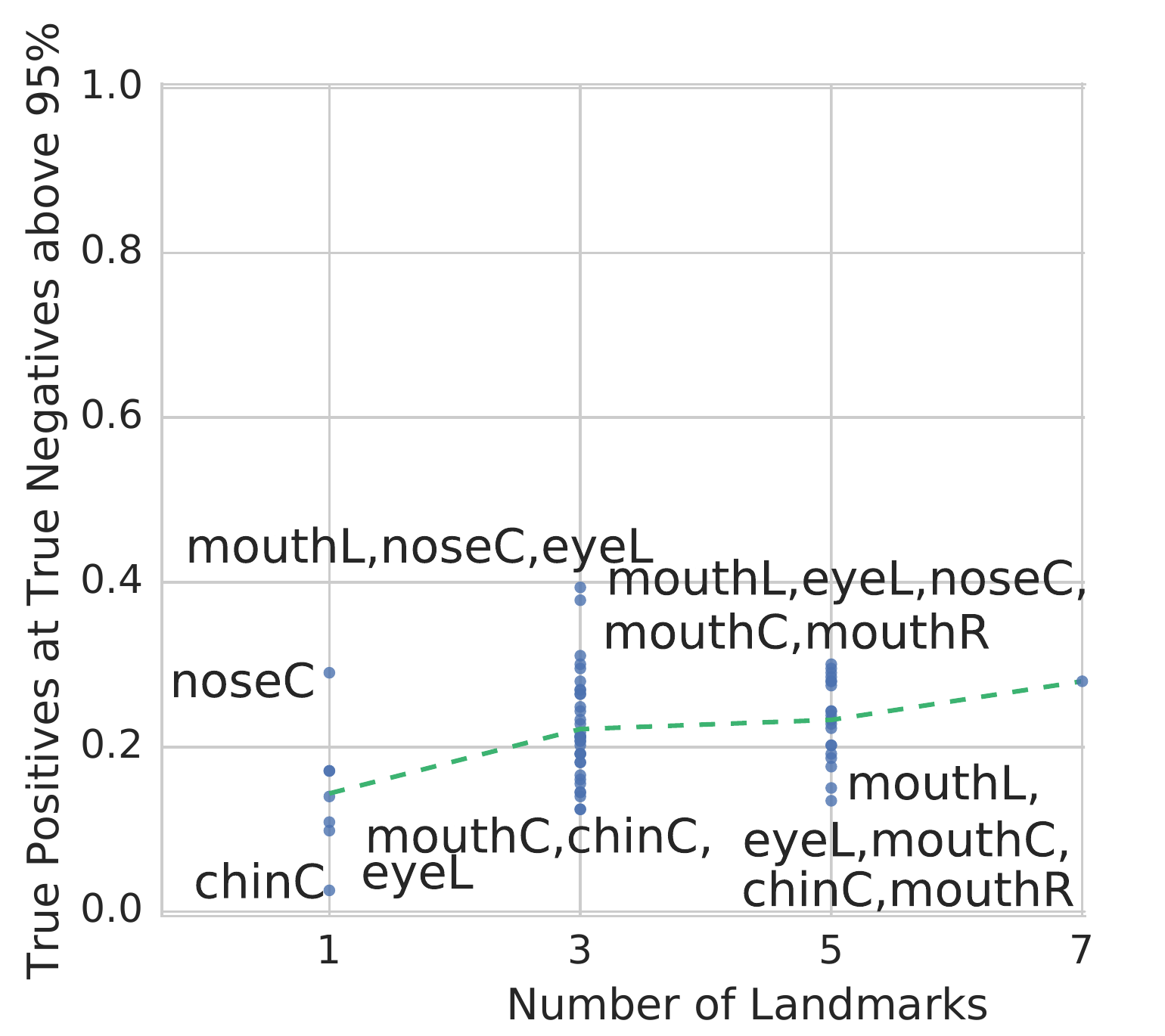}
    \caption{\label{fig:AFLW-uricar-scatter_plot}{\tiny Uricar AFLW}}
  \end{subfigure}%
  \begin{subfigure}[b]{0.25\linewidth}
    \centering\includegraphics[width=\linewidth]{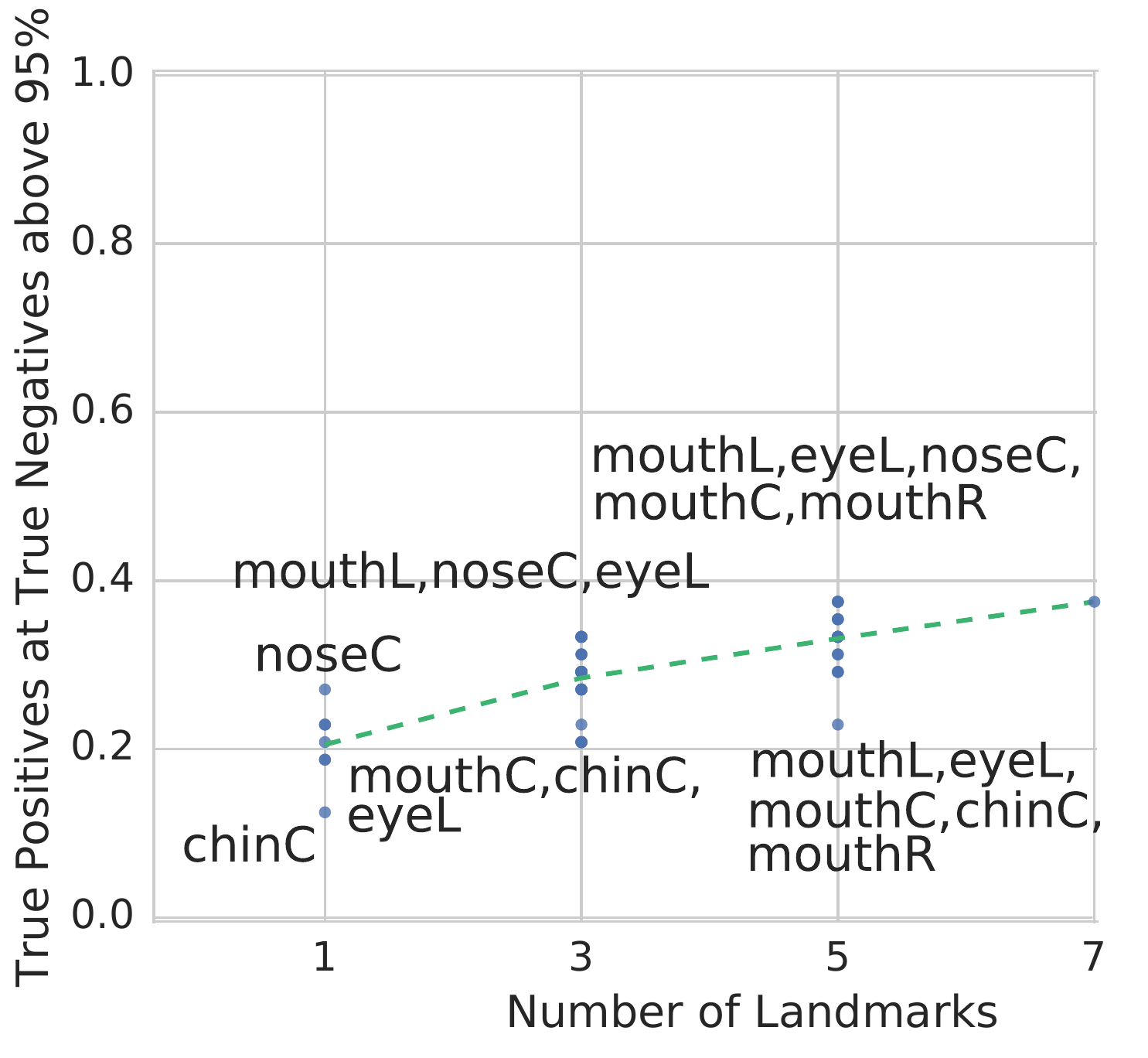}
    \caption{\label{fig:AFLW-kazemi-scatter_plot}{\tiny Kazemi AFLW}}
  \end{subfigure}%
  \begin{subfigure}[b]{0.25\linewidth}
    \centering\includegraphics[width=\linewidth]{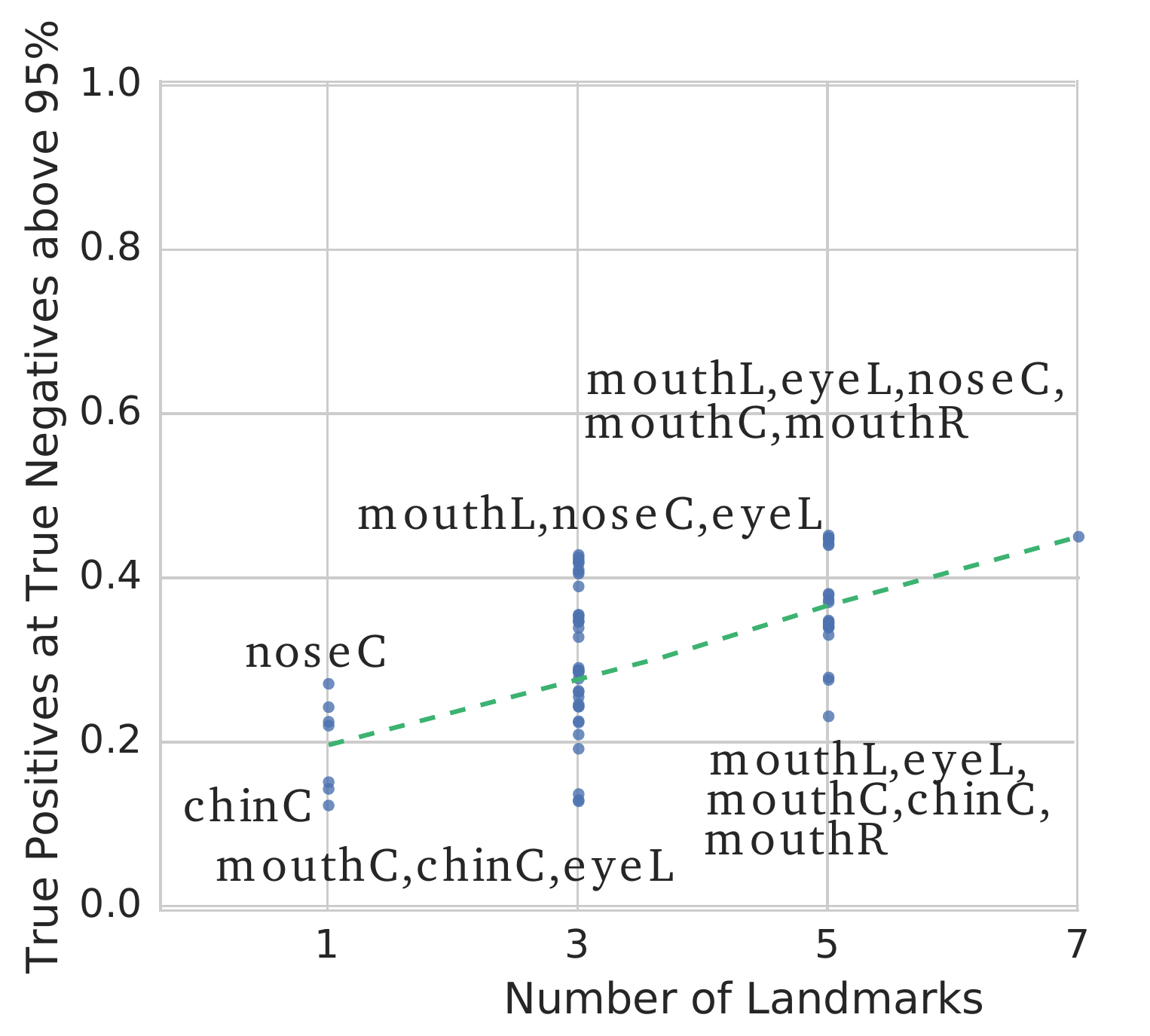}
    \caption{\label{fig:AFLW-gen-sec-scatter_plot}{\tiny Gen. AFLW Casc.}}
  \end{subfigure}
  
  \begin{subfigure}[b]{0.25\linewidth}
    \centering\includegraphics[width=\linewidth]{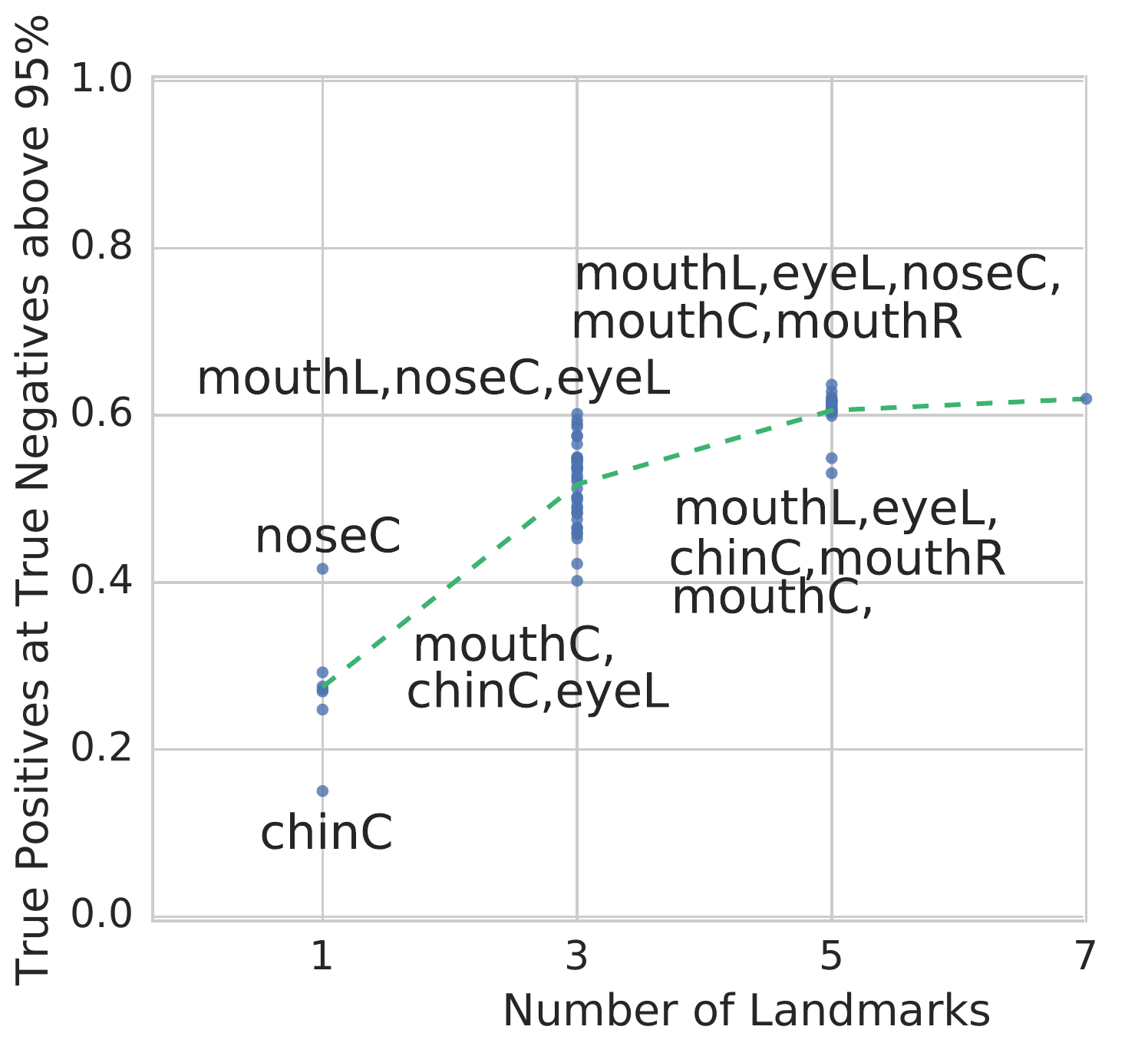}
    \caption{\label{fig:HELEN-gen-scatter_plot}{\tiny Gen. HELEN}}
  \end{subfigure}%
  \begin{subfigure}[b]{0.25\linewidth}
    \centering\includegraphics[width=\linewidth]{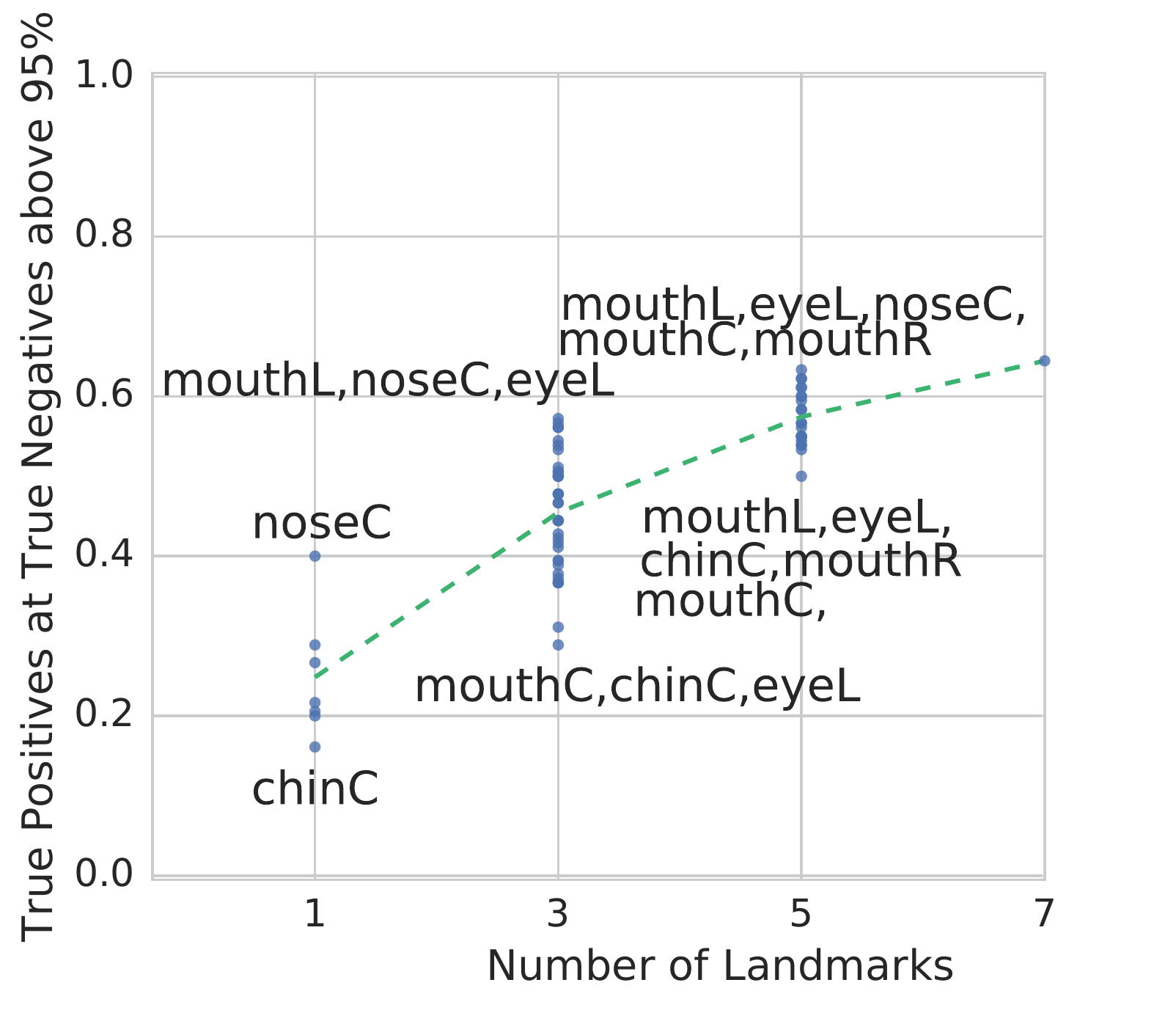}
    \caption{\label{fig:HELEN-uricar-scatter_plot}{\tiny Uricar HELEN}}
  \end{subfigure}%
  \begin{subfigure}[b]{0.25\linewidth}
    \centering\includegraphics[width=\linewidth]{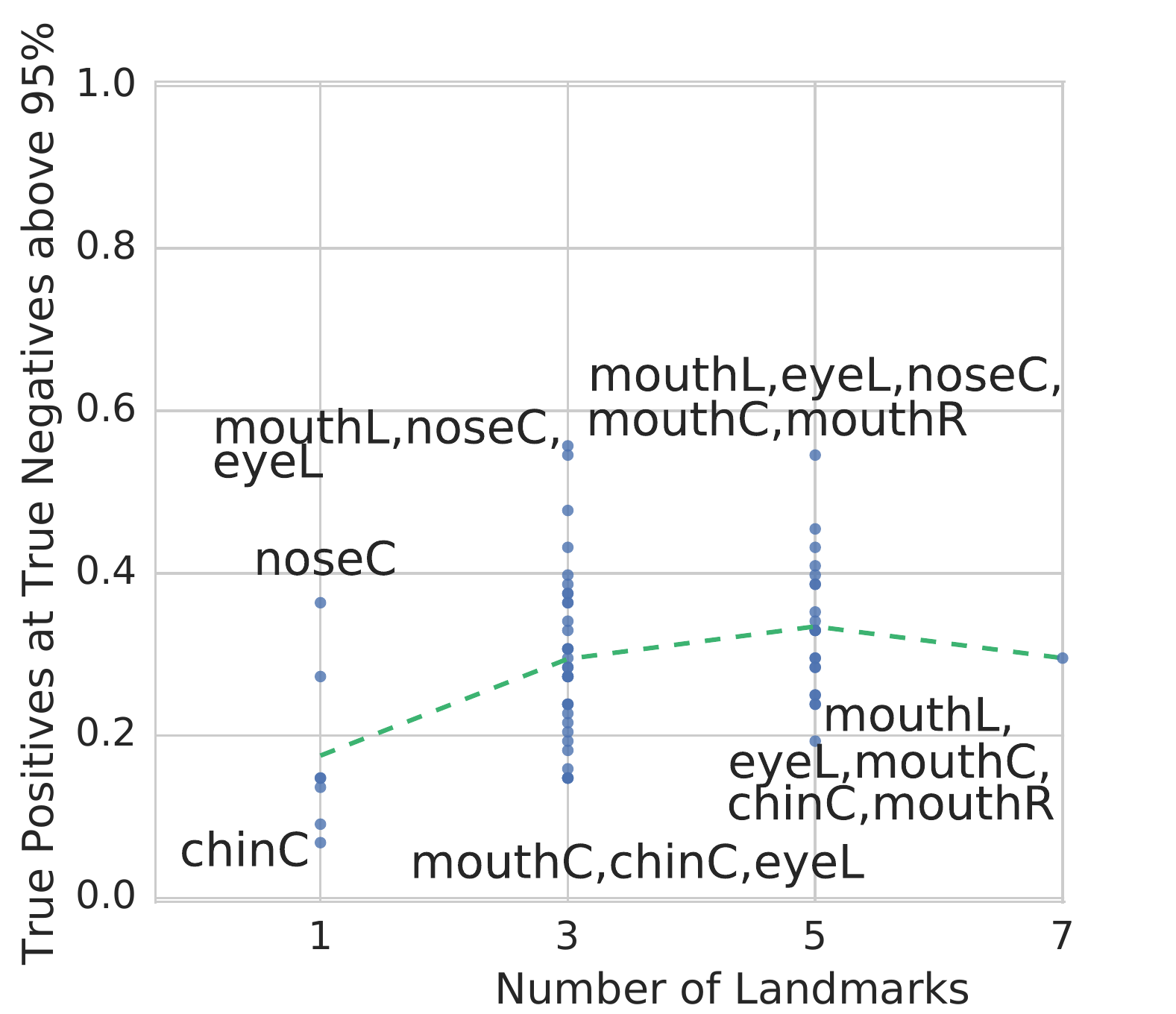}
    \caption{\label{fig:HELEN-kazemi-scatter_plot}{\tiny Kazemi HELEN}}
  \end{subfigure}%
  \begin{subfigure}[b]{0.25\linewidth}
    \centering\includegraphics[width=\linewidth]{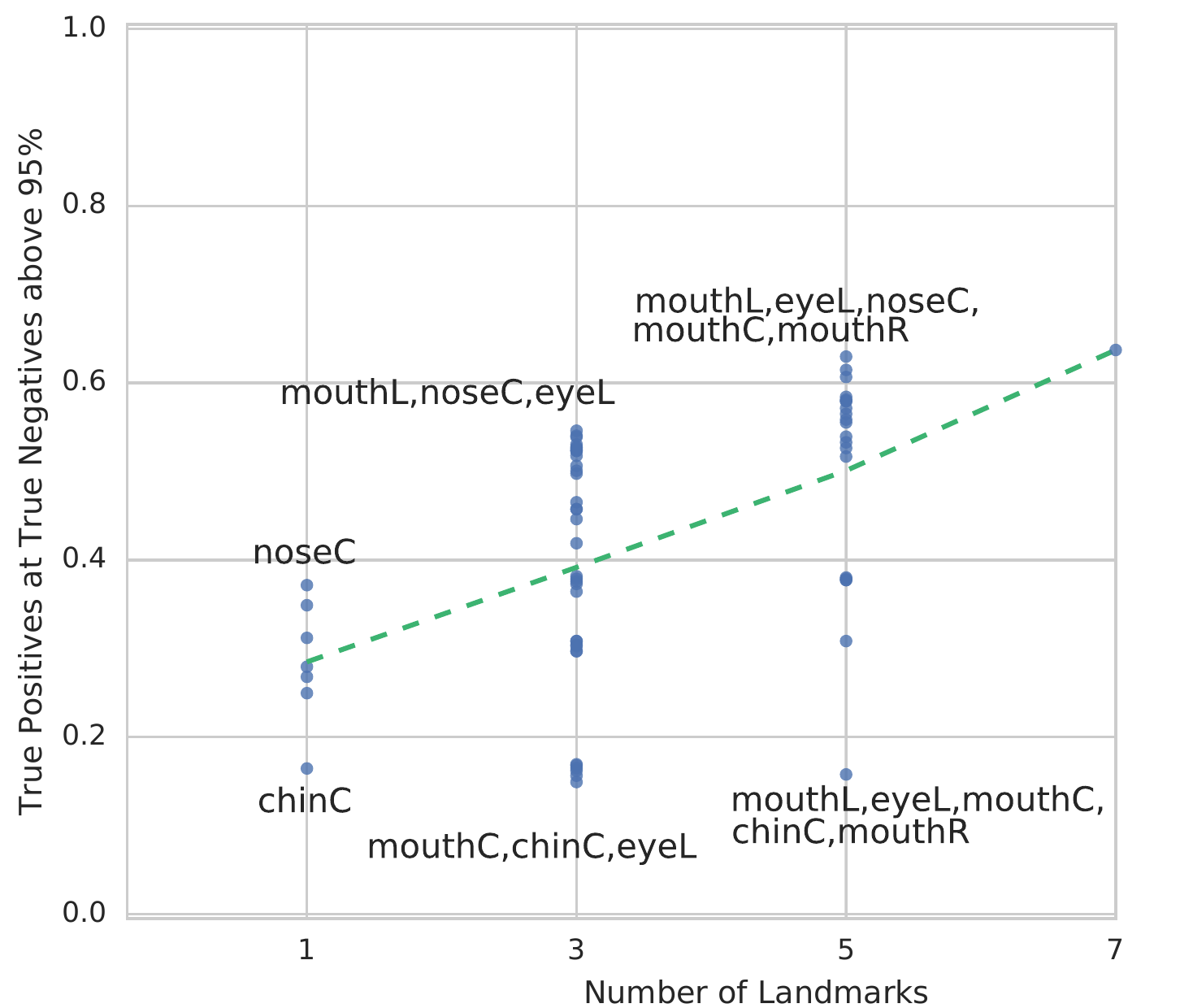}
    \caption{\label{fig:HELEN-gen-sec-scatter_plot}{\tiny Gen. HELEN Casc.}}
  \end{subfigure}
  \caption{\label{fig:scatter}
Rate of Failures that are correctly marked as failures at an operating point where the rate of Non Failures that are correctly marked as Non Failures is above 95\% for different combinations of landmarks. The dashed line represents the mean for each number of landmarks. Those with caption ``Casc.'' were computed with the Cascaded SVR model the others with the Joint SVR. ``Gen.'' stands for Generated Landmarks.
  }
\end{figure}

\textbf{Joint Features} (Figure~\ref{fig:svm_config}a) For that method we concatenated the feature vectors of different landmarks. Again we used the SVR with the same properties as in Section~\ref{ssc:confidence_landmark}. Figure~\ref{fig:scatter} shows the accuracy of different combinations of landmarks on the validation set that are used to extract features for training the predictor. For training we used 10000 samples and the best feature combination (Section \ref{ssc:confidence_landmark}). In order to tune the threshold for the predicted confidence C for the TrueCorrect95 we use a disjunct partition of the validation set and tuned it according to Figure~\ref{fig:tune_threshold_lm}. Figure~\ref{fig:tune_precision_lm} shows the result for different thresholds for the ground truth confidence C. The vertical tick shows the prior chosen threshold that is used to be able to compare Uricar~\cite{uricar2015real} and Kazemi~\cite{kazemi2014one} Detector prediction.
 
\textbf{Cascaded SVRs}
In our second setup we used the predicted Confidence C of each individual landmark from the SVR (Section~\ref{ssc:confidence_landmark}) as features for a second SVR (Figure~\ref{fig:svm_config}b). Therefore we predicted the confidence C on the test and validation set that we chose from AFLW and HELEN (Section~\ref{ssc:datasets}). The second SVR was then trained on the validation set because the errors of the predicted confidence C from the training set would be overoptimistic. For the training the same parameters as in Section~\ref{ssc:confidence_landmark} and 3200 samples were used. 

For almost all experiments the accuracy increases with number of landmarks that are used for training. The combination of nose, left mouth corner and left eye is already very close to the best combination in terms of TrueCorrect95. The most informative landmark is the nose, which is contained in all sets of landmarks with the best prediction. The least informative is the chin which is contained in all sets of worst performing configurations. The proposed method predicts more accurate results for the HELEN dataset than on AFLW.

\section{Application}
\label{sec:application}
In the following we apply our failure prediction approach to gender estimation from face images, a typical application from computer vision. First we describe the application pipeline and define a fast method and a slow robust method for landmarks detection, then we describe the gender prediction and analyze the trade off between fast and robust methods based on our predicted confidence for the landmark detections.
\subsection{Pipeline}
\label{ssc:pipelie}

\begin{figure}[htb!]
\centering
\includegraphics[width=0.7\textwidth]{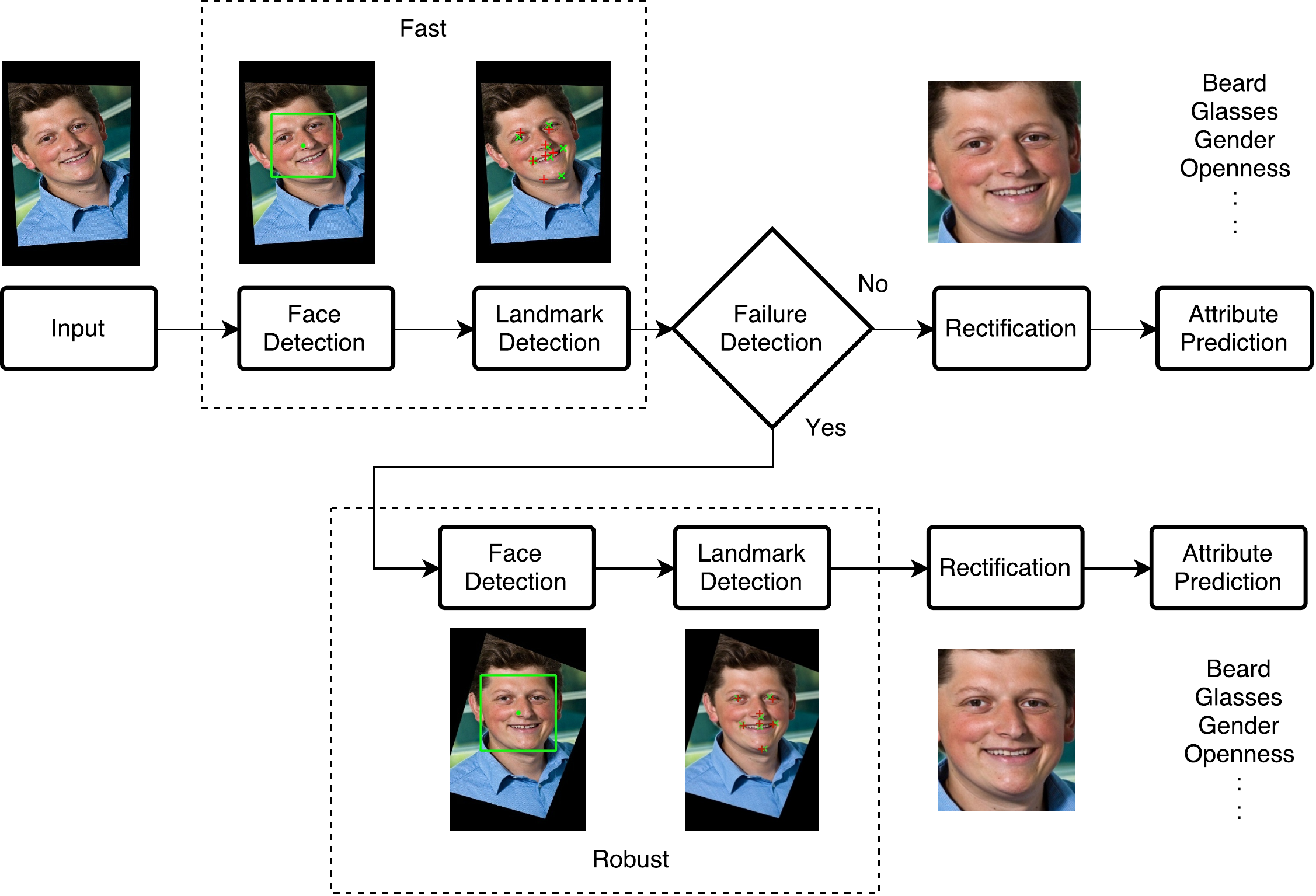}
\caption{
Pipeline of Handling failures for gender estimation. 
}
\label{fig:gender_pipeline}
\vspace{-0.5cm}
\end{figure}

Our pipeline of how we predict the gender of a person is depicted in Figure~\ref{fig:gender_pipeline}. In order to make the prediction more robust we distorted the images from AFLW~\cite{Kostinger-ICCVW-2011}. In this step we generated 5 new images from one in the AFLW dataset. Moreover we only used images with yaw and pitch smaller than 15 degree and roll smaller than 60 degree. This step is depicted in Figure~\ref{fig:gender_pipeline} (a) to (b). Therefore we used following distortions:
\begin{enumerate}
	\item For perspective transformation we used a Gaussian distributed ratio for the new length for both top width and the right height of the image. We used a standard  of 0.05 and a mean of 1 for the scaling factor.
    \item For rotation we used a standard deviation of 10.
    \item For Gaussian noise in the pixel values we used a mean of 10 and a standard deviation of 5.
\end{enumerate}

The next step is to detect the face. To do so we used the openCV face detector with the Haar feature-based cascade model that returns a rectangular in which the face is detected and a score for the goodness of the detection. Because most faces are already rotated properly we penalize detections with large angles. Moreover we filter detections of faces that are very small compared to the image size. 

\textbf{Fast Method} In order to detect the faces we first rotate each image by 45 degree from -90 to 90 degrees. The runtime per image is 2.67s. Then we apply the face detection to each image and choose the highest scored.

The Uricar Detector~\cite{uricar2015real} is then used to predict the landmarks of the detected face. In that step we apply our failure detection that was train for a combination for all landmarks according to Section~\ref{ssc:confidence_face}. The face detection for this method takes 147ms per image. For all faces with landmarks that were predicted as being wrong we use the slow robust method to predict new landmarks.

\textbf{Slow Robust Method} For the robust method we fall back to the face detection from openCV. To get a better aligned image we rotate it before the detection in steps of 5 degree from -90 to 90 degrees. The runtime per image is 20.1s. On this new aligned face detection that we got from the highest scored image, we apply the landmark detector.

\subsection{Gender Prediction}
\label{ssc:gender_prediction}

From the landmarks obtained by either the fast or slow robust method we extract the features for the gender prediction. For that we calculate a grid of SIFT and LBP descriptors of the landmarks eye left, eye right, nose center, mouth left, mouth center, mouth right and chin center as described in Figure \ref{fig:landmark_names} (d). Moreover a Grid of SIFT descriptors over the whole face. As classifier we use the SVC from scikit-learn~\cite{scikit-learn}. We use 5 fold cross-validation to determine the best parameter from a exhaustive search over $C \in \{0.1, 0.3, 0.5, 0.7\}$ and the kernels $\in \{$linear, poly (with degree = 3), RBF and sigmoid$\}$. 

\subsection{Trade off between Fast and Robust}
\label{ssc:trade_off}

\begin{figure}[th!]
  \centering
  \begin{subfigure}[b]{0.46\linewidth}
    \centering\includegraphics[width=\linewidth]{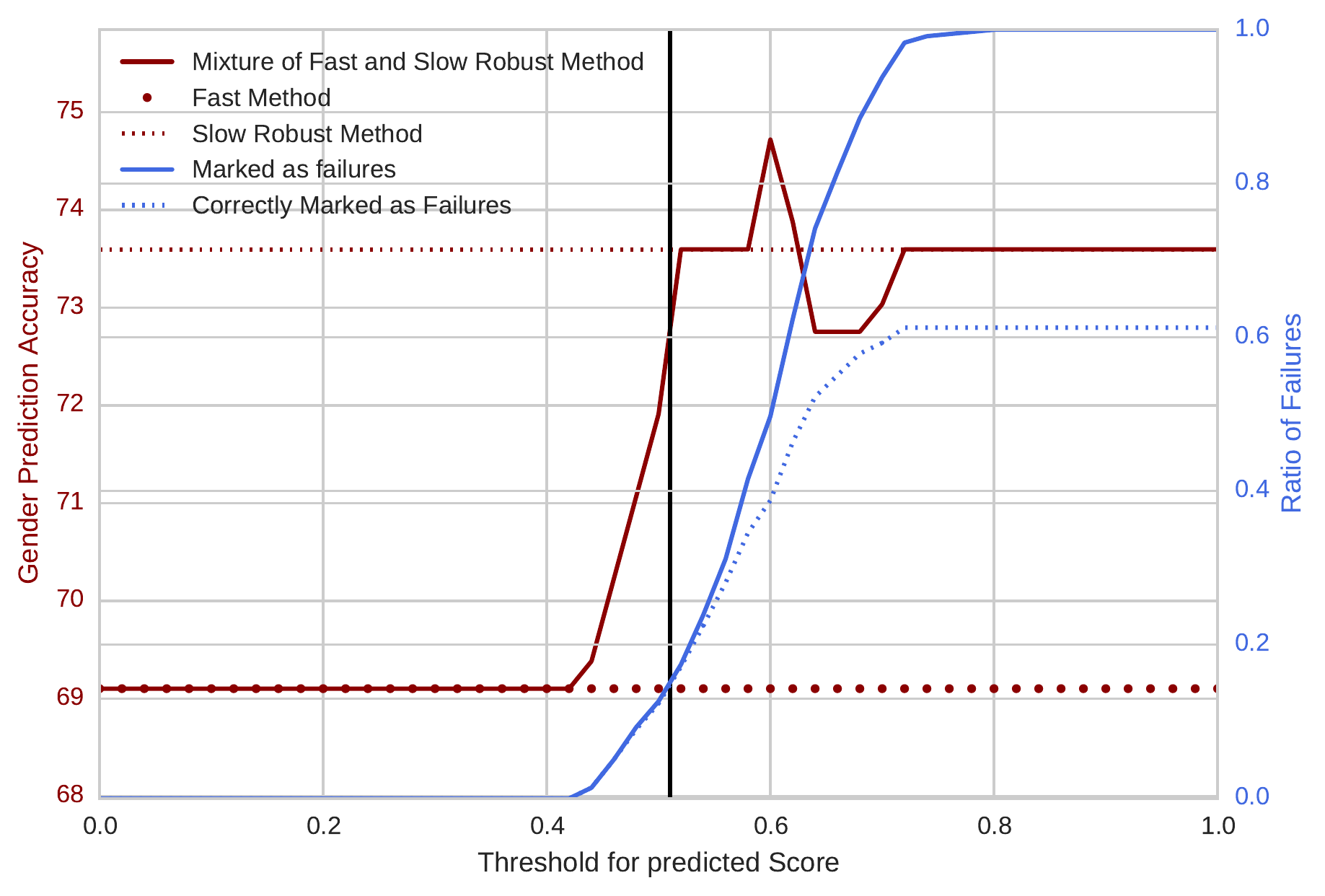}
    \caption{\label{fig:gen_plt}\small Accuracy of Gender Predictor}
  \end{subfigure}%
  \begin{subfigure}[b]{0.46\linewidth}
    \centering\includegraphics[width=\linewidth]{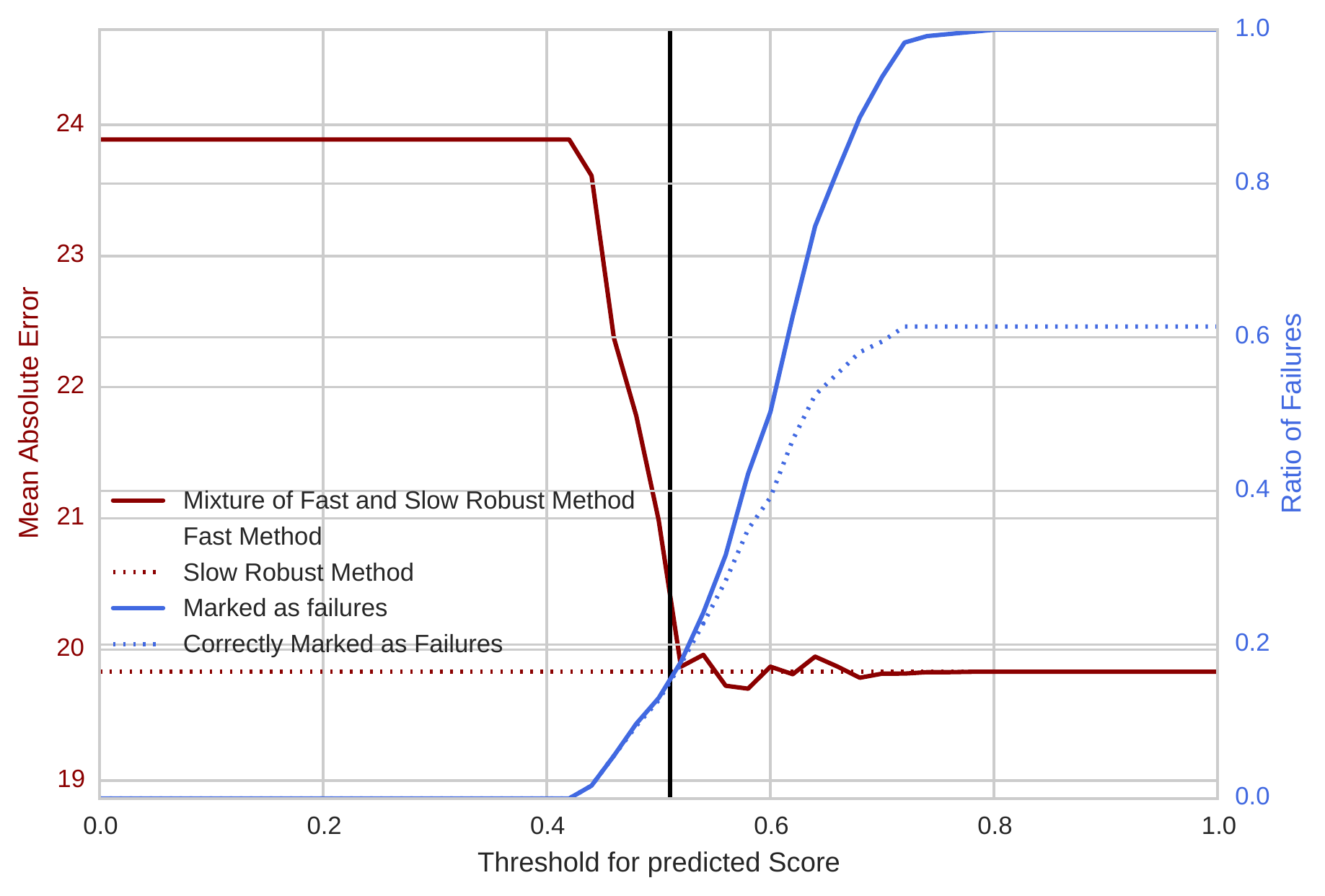}
    \caption{\label{fig:acc_plt}\small Error of Landmark predictor}
  \end{subfigure}%
  \caption{\label{fig:trade_off_robust}
Trade off between fast method and slow robust method based on predicted Confidence C of each face for Uricar Detector on AFLW.
  }
    \vspace{-0.75cm}
\end{figure}

Figure~\ref{fig:trade_off_robust} shows the trade off between the fast and robust method over a range of thresholds for the predicted confidence. All faces that were detected as failed detection are recalculated with the robust method. The threshold for the ground truth confidence was fixed to 0.65 and the Confidence C was calculated with a standard deviation of 10\%. The plots are reported on the test set and all parameters were tuned on a disjunct set.

Figure~\ref{fig:gen_plt} shows that the gender predictor accuracy can almost reach the accuracy of the slow robust method when only recalculating 15.4\% of the face images. The threshold for the predicted Confidence C is at 0.51, which was chosen during training by using the TrueCorrect95. The fast method which takes 2.92s per image has an accuracy of 69.1\% and the slow robust method which takes 20.3s per image an accuracy of 73.6\%. The trade off between those two takes 6.05s per image at chosen operating point and has a gender prediction accuracy of 72.8\%. Therefore we reach about the same precision as the robust method in 70\% less time or $3.36\times$ faster.

Figure~\ref{fig:acc_plt} shows that the MAE can also almost reach the accuracy of the slow robust method when only recomputing 15.4\%. The MAE of the fast method is 23.9px, of the robust method is 19.8px and of the trade off between those two at the operating point is 20.4px.
 
\section{Conclusion}
\label{sec:conclusion}
This paper proposed a novel method to predict the confidence of detected landmarks. To train the model we used the error in distance as measure for the confidence of a landmark. With this method it is possible to decide in a trade off between number of detected failures and the accuracy of detecting failures. Meaning that if we want to have less correct face detections detected as failed, we can use a higher confidence threshold to detect only those as failure that are more likely to be failed. To validate our results we used two popular face image data sets and two landmark detectors. To demonstrate the benefit of our method we showed that we can improve the speed of a gender predictor pipeline by adding a fast method. It falls back to a slow robust method if the face is detected as being failed. That gave us a speed up of $\times3$ while achieving almost the same accuracy as the robust method.

\clearpage

\bibliographystyle{splncs}
\bibliography{egbib}

\end{document}